\documentclass{article}

\usepackage{microtype}
\usepackage{graphicx}
\usepackage{subcaption}
\usepackage{booktabs}
\PassOptionsToPackage{hyphens}{url}
\usepackage{hyperref}

\usepackage[accepted]{icml2026}

\usepackage{amsmath}
\usepackage{amsfonts}
\usepackage{amssymb}
\usepackage{xcolor}
\usepackage{float}
\usepackage{tabularx}
\usepackage{multirow}
\usepackage{url}

\definecolor{darkblue}{rgb}{0, 0, 0.5}
\hypersetup{colorlinks=true, citecolor=darkblue, linkcolor=darkblue, urlcolor=darkblue}

\icmltitlerunning{Where's the Plan? Locating Latent Planning in Language Models with Lightweight Mechanistic Interventions}

\begin{document}

\twocolumn[
  \icmltitle{Where's the Plan? Locating Latent Planning in Language Models with Lightweight Mechanistic Interventions}

  \icmlsetsymbol{equal}{*}
  \icmlsetsymbol{stanford}{1}

  \begin{icmlauthorlist}
    \icmlauthor{Nicole Ma}{equal,stanford}
    \icmlauthor{Nick Rui}{equal,stanford}
  \end{icmlauthorlist}

  \icmlkeywords{mechanistic interpretability, latent planning, language models, activation patching, probing}

  \vskip 0.3in
]

\makeatletter
\gdef\icmlcorrespondingauthor@text{\{manicole, nickrui\}@stanford.edu}
\makeatother
\printAffiliationsAndNotice{\textsuperscript{*}Equal contribution. Authors listed by increasing height. \textsuperscript{1}Stanford University}

\begin{abstract}
We study \textit{planning site formation} in language models---\textit{where} internal representations of structurally-constrained future tokens form during the forward pass, and whether they causally drive generation. Using rhyming-couplet completion as a clean test of forward-looking constraint, we apply two lightweight methods (linear probing and activation patching) across Qwen3, Gemma-3, and Llama-3 at more than ten scales. Probing shows that future-rhyme information is linearly decodable at the line boundary, with signal that strengthens with scale in all three families. Activation patching reveals that only Gemma-3-27B causally relies on this encoding, exhibiting a \textit{handoff} in which the causal driver migrates from the rhyme word to the line boundary around layer 30. Every other model we test conditions on the rhyme word throughout generation, with near-zero causal effect at the line boundary despite strong probe signal. We localize the Gemma-3-27B handoff to five attention heads through two-stage path patching that recover ${\sim}90\%$ of the rhyme-routing capacity at the newline.
\end{abstract}

\section{Introduction}

Autoregressive language models generate text one token at a time, yet routinely produce outputs requiring long-range structural coherence. Rhyming couplets, for example, require that tokens generated late in the sequence stand in a precise phonological relation to a stimulus word from much earlier. This raises a natural question: do models form internal representations of future outputs that causally shape generation, entirely invisible to behavioral evaluation? We call this \textit{latent planning}. Unlike chain-of-thought reasoning~\citep{wei2022chain}, where intermediate steps are directly observable, latent planning occurs entirely within a model's hidden activations. If models are planning in ways invisible to external observers, standard behavioral evaluations may systematically underestimate this capacity, making it both a scientific and a safety-relevant problem~\citep{pfau2024dots,hao2024coconut}.

Neural networks trained on next-step prediction can develop internal planning representations in structured domains. McGrath et al.~\citeyearpar{mcgrath22} find that chess concepts emerge in AlphaZero's internal layers; Li et al.~\citeyearpar{li24} show that transformers trained on Othello develop board-state representations, and Nanda et al.~\citeyearpar{nanda23} show that these representations are linear and causally manipulable; and Jenner et al.~\citeyearpar{jenner24} provide mechanistic evidence of learned look-ahead in a chess-playing network. These results motivate the question of whether language models elicit analogous mechanisms during open-ended generation. Prior work on large language models establishes that planning-compatible information exists in some models~\citep{maar2026whatsplanmetricsimplicit,hanna2026latent,dong2025emergentresponseplanningllms,pochinkov2025parascopeslanguagemodelsactivations}, but does not address a more specific question: \textit{where exactly does planning information reside during the forward pass, and does it move?} We call this \textit{planning site formation}.

Investigating planning sites rigorously requires both encoding evidence (what information is present) and causal evidence (what information is used). Probes assess what is encoded in hidden states~\citep{hewitt19,burns24}, but sufficiently flexible probes can achieve high accuracy by memorizing labels rather than reflecting genuine representations~\citep{hewitt19}. Causal tools such as activation patching~\citep{meng23,wang2022ioi} and steering vectors~\citep{turner2023activation,arditi2024refusal} are needed to establish that encoded information actually influences downstream generation. The most expressive existing approach, training transcoders~\citep{transcoders} to build feature circuits~\citep{sparsefeaturecircuits,ameisen2025circuittracing}, provides fine-grained circuit analysis but requires substantial compute (effectively a second training run), has so far been applied only to closed-source models such as Claude 3.5 Haiku~\citep{anthropic-bio}, and does not straightforwardly scale to new open-source architectures. Work by \citet{maar2026whatsplanmetricsimplicit} uses steering vector interventions and finds most open-source models up to 30B parameters keep their planning sites at the last word token; a transcoder-based analysis by \citet{anthropic-bio} on Claude 3.5 Haiku finds evidence of a planning site migrating to the newline. These results motivate a scalable framework for studying planning site formation across model architectures and scales.

In this paper, we define two notions for surfacing evidence of latent planning: the weaker notion of planning-compatible representations (detectable via probing) and the stricter notion of causally active planning sites (established via activation patching). Using only linear probing and activation patching (requiring no transcoder training and far less data than steering vector approaches), we investigate planning site formation across three open-source model families at multiple scales (up to 70B parameters).

Probing reveals that planning-compatible representations emerge at the newline token with scale across all three families, yet their strength and layer profile vary substantially across architectures. Activation patching reveals that only Gemma-3-27B forms a causally active planning site at the newline, exhibiting an information routing handoff in which causal influence migrates from the last word token to the newline around layer 30. All other models, including all Qwen3 sizes up to 32B and all Llama-3 sizes up to 70B, condition on the last word token throughout generation despite encoding planning-compatible representations at the newline. We further localize the handoff in Gemma-3-27B to a sparse set of five attention heads in layers 28 and 30, identified via attention weight ranking and simultaneous head patching. This suggests that planning site formation is a distinct emergent phenomenon tied to specific model scale and architecture, rather than a general consequence of strong planning-compatible representations.

\section{Setup and Notation}
\label{sec:2}
In this paper, we study rhyming couplet generation as an example of latent planning.
A rhyming couplet is a two-line poem where the last word of the first line, denoted $r_1$, rhymes with the last word of the second line, denoted $r_2$.
We task language models to complete rhyming couplets (i.e., given context containing $r_1$, generate a completion where $r_2$ rhymes with $r_1$). For example, given the prompt \texttt{"A rhyming couplet:\textbackslash nShe felt a sudden sense of fright,\textbackslash n"}, the model should produce a completion such as \texttt{"and hoped that dawn would end the night.\textbackslash n"}, where $r_1 = \texttt{fright}$ and $r_2 = \texttt{night}$.

Consider an autoregressive transformer language model with $L$ layers and hidden dimension $d$.
Let $\mathbf{h}_{\ell,i} \in \mathbb{R}^d$ denote the hidden state vector after the $\ell$-th transformer block at position $i$.
We adopt a relative positioning scheme centered around the newline token $\verb|\n|$ ending the first line of the couplet.
We refer to this token as being at position 0, with position $i$ denoting the token $i$ steps before (negative) or after (positive) the newline.

We say $(i, \ell)$ contains a \textit{planning-compatible representation} if the generated $r_2$ can be decoded from $\mathbf{h}_{\ell,i}$ via probing substantially better at position $i$ than at other positions.
This signals that information about the future token or rhyme scheme is disproportionately encoded at $(i, \ell)$ relative to other hidden states.
As a stricter definition, we say $(i, \ell)$ is a \textit{causally active planning site} if replacing $\mathbf{h}_{\ell,i}$ during generation with the hidden state from a run targeting a different rhyme substantially redirects output toward that rhyme.
This shows that rhyme information located at $(i, \ell)$ is causally used in generation.

\section{Probing for Planning-Compatible Representations}
\label{sec:probing}
To investigate planning-compatible representations, we train linear probes to predict future tokens from hidden states.
Let $\mathcal V$ be the vocabulary (set of all tokens) and $\Delta \mathcal V$ the probability simplex over $\mathcal V$.
Each linear probe is a parameterized function $f_{(W,b)}(\mathbf{h}): \mathbb{R}^d \to \Delta \mathcal V$ where $f_{(W,b)}(\mathbf{h}) = \text{softmax}(W\mathbf{h} + \mathbf{b})$.
Probes are trained by minimizing cross-entropy loss. Probe parameters are optimized with AdamW~\citep{adamw} with learning rate $10^{-4}$, weight decay $10^{-3}$, batch size 32, for 10 epochs. We report Wilson 95\% confidence intervals~\citep{wilson1927} on every probe accuracy, computed from the size of the held-out evaluation set.

\subsection{Probing general text as a negative control}
\label{sec:probe-general}

We first verify that planning-compatible representations are task-specific and do not occur in general text generation.
We randomly sample 1,200 (1,000 train, 200 validation) token sequences from The Pile~\citep{thepile}, greedily sample model completions while storing activations to build the probe training dataset.
For each layer $\ell$ and various look-ahead distances $k$, we train a linear probe to predict the generated token at position $i+k$ from $\mathbf{h}_{\ell,i}$.
We compare against a baseline unigram model (trained on the corpus of all model completions) to ensure probes are learning more than just token frequencies.

\begin{figure*}[t]
    \centering
    \begin{subfigure}{0.325\linewidth}
        \centering
        \subcaption[]{Qwen3-32B}
        \includegraphics[width=\linewidth]{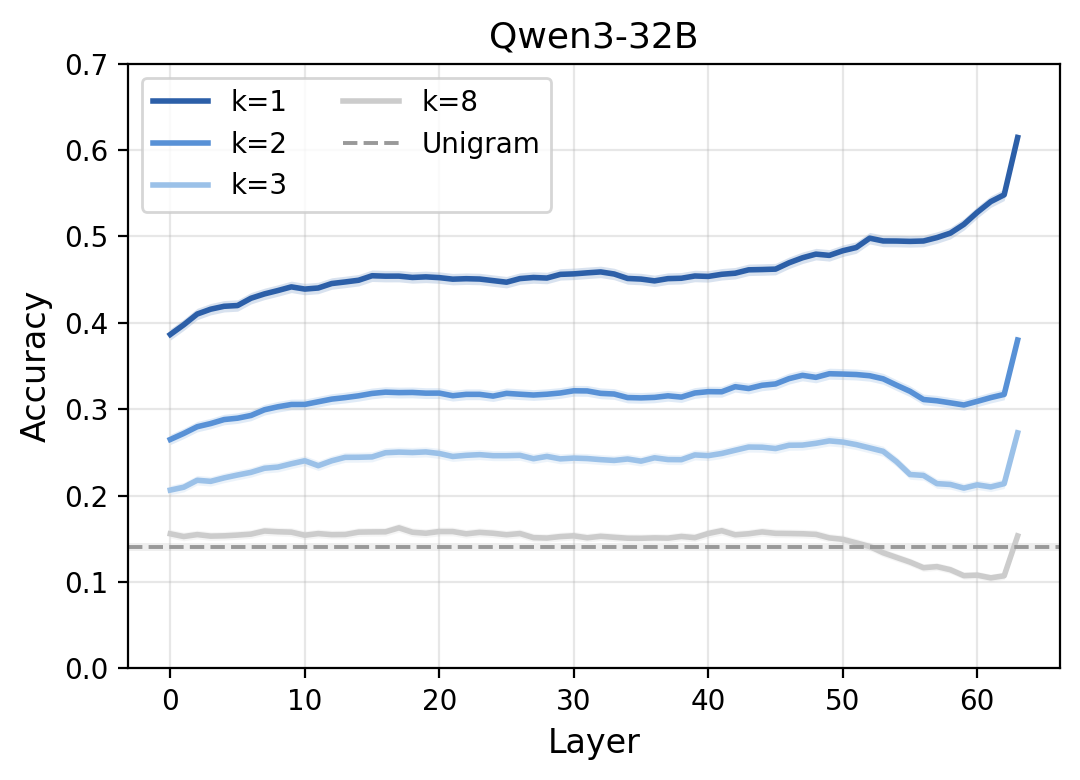}
    \end{subfigure}
    \begin{subfigure}{0.325\linewidth}
        \centering
        \subcaption[]{Gemma-3-27B}
        \includegraphics[width=\linewidth]{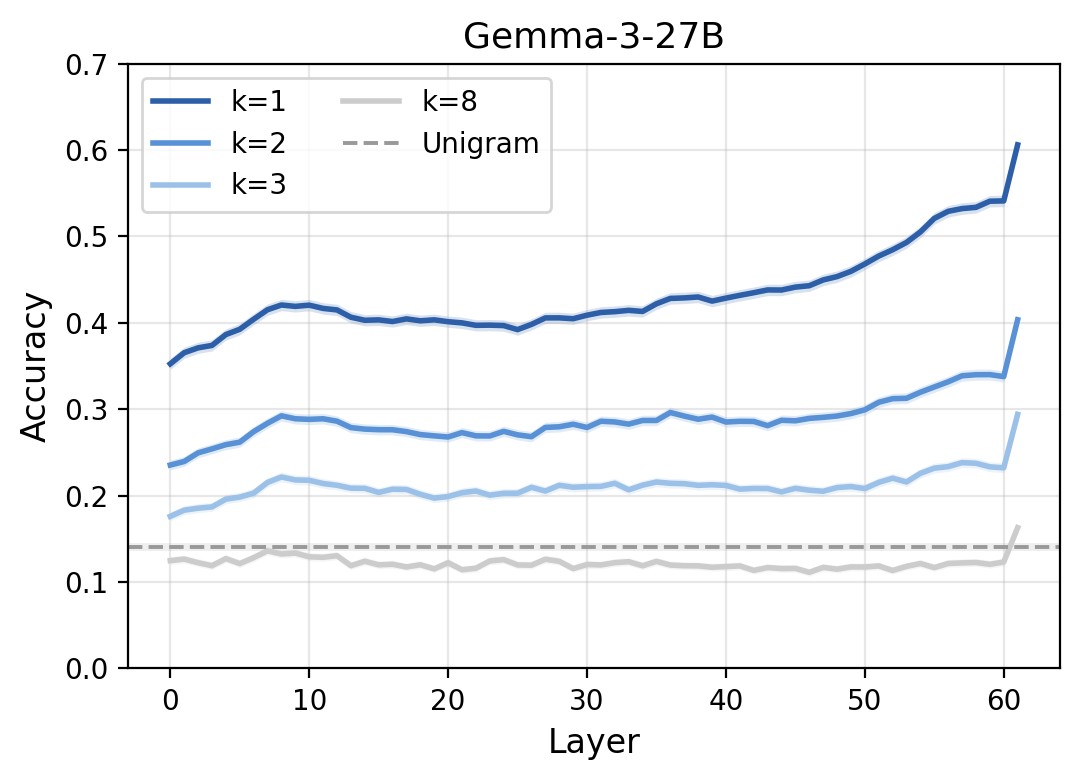}
    \end{subfigure}
    \begin{subfigure}{0.325\linewidth}
        \centering
        \subcaption[]{Llama-3.1-70B}
        \includegraphics[width=\linewidth]{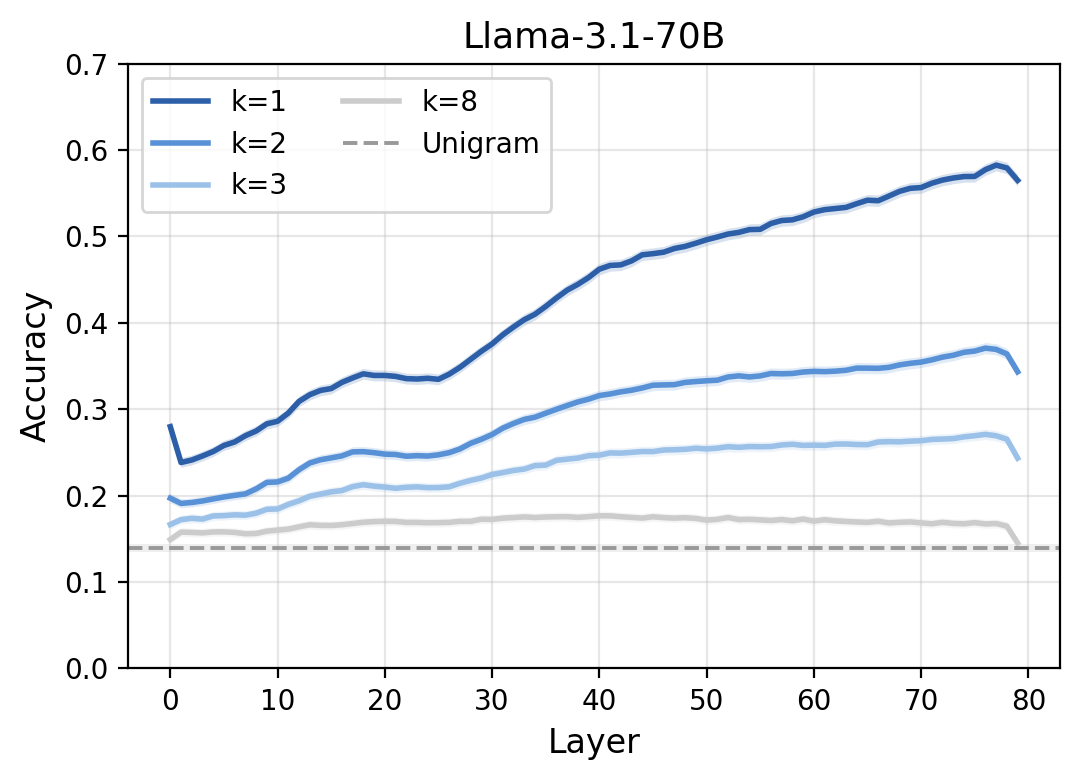}
    \end{subfigure}
    \caption{Top-5 accuracy of linear probes predicting $k$ tokens ahead in general text (Pile). Wilson 95\% CI bands are drawn but visually imperceptible: per-token $N{\approx}21{,}000$ gives a half-width of ${\sim}0.005$ at typical $p$, so the curves are essentially noise-free at this sample size. Accuracy degrades monotonically with $k$ and the $k=8$ curve overlaps the unigram baseline across all layers, confirming that planning-compatible representations are not a generic feature of the residual stream.}
    \label{fig:probe-results}
\end{figure*}

We find probe accuracy degrades monotonically with $k$ and the $k=8$ confidence band overlaps the unigram baseline across all layers in all three models (Figure~\ref{fig:probe-results}), confirming that planning-compatible representations are not a generic property of the residual stream. This provides a negative control establishing that probe signal in Section~\ref{sec:probe-couplets} is significant.

\subsection{Probing rhyming couplets}
\label{sec:probe-couplets}
\begin{figure}[t]
    \centering
    \includegraphics[width=\linewidth]{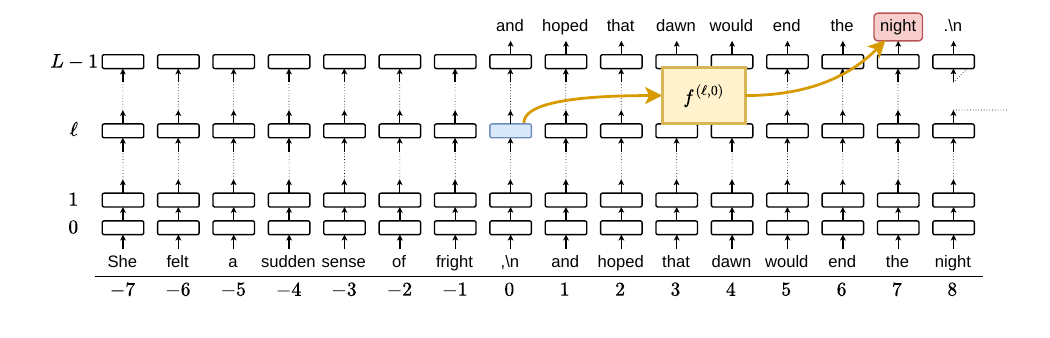}
    \caption{A linear probe $f^{(\ell,0)}$ trained to predict the rhyming token from
    activations at the newline position ($i=0$).}
    \label{fig:poem-diagram}
\end{figure}

For rhyming couplet generation, we first synthetically generate 1,200 (1,000 train, 200 validation) rhyming couplets with Claude Sonnet 4.6 \citep{sonnet46}, strategically prompting for diversity of topics and rhyme schemes.
We truncate the second line of each couplet, then greedily sample model completions for the second line, storing activations to build the probe training dataset.
We train probes to predict the rhyming token $r_2$ from $\mathbf{h}_{\ell,i}$ (Figure~\ref{fig:poem-diagram}), evaluating on raw accuracy and rhyme accuracy (as determined by the CMU Pronouncing Dictionary \citep{cmu}) across layers and positions.
The comparison is between probes trained on activations at the newline position ($i=0$) and subsequent positions ($i > 0$).
If a model constructs a planning-compatible representation at the newline before generation begins, the $i=0$ probe should substantially outperform $i>0$ probes. We also probe on hidden states at the last word token position,\footnote{Position $i=-2$ in Gemma and $i=-1$ in Qwen and Llama (see Appendix~\ref{sec:model-config}).} which we expect to perform well since the last word directly decides the rhyme scheme of the couplet.

\begin{figure*}[t]
    \centering
    \begin{subfigure}{0.325\linewidth}
        \centering
        \subcaption[]{Qwen3-32B (Top-5 Acc.)}
        \includegraphics[width=\linewidth]{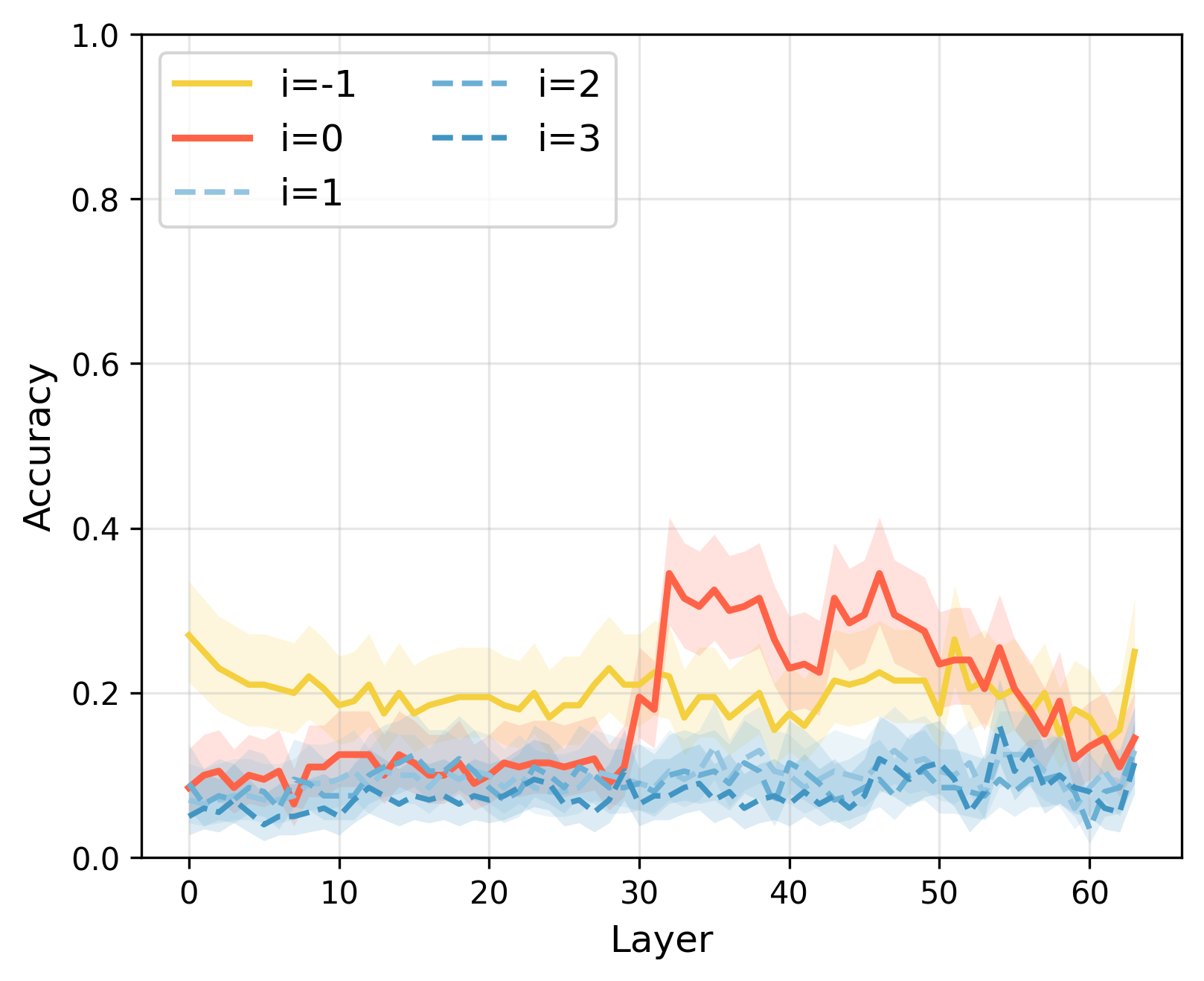}
    \end{subfigure}
    \begin{subfigure}{0.325\linewidth}
        \centering
        \subcaption[]{Gemma-3-27B (Top-5 Acc.)}
        \includegraphics[width=\linewidth]{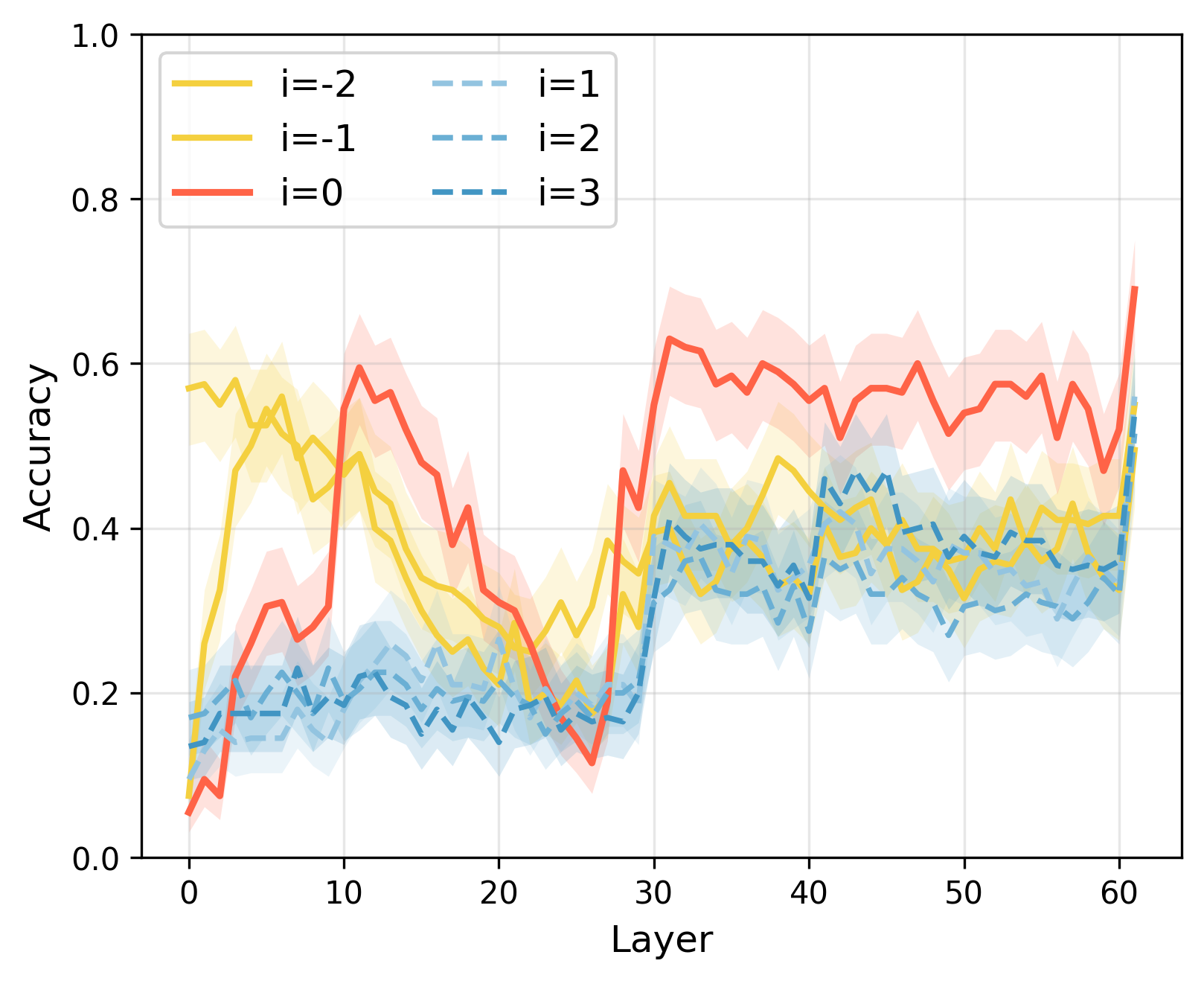}
    \end{subfigure}
    \begin{subfigure}{0.325\linewidth}
        \centering
        \subcaption[]{Llama-3.1-70B (Top-5 Acc.)}
        \includegraphics[width=\linewidth]{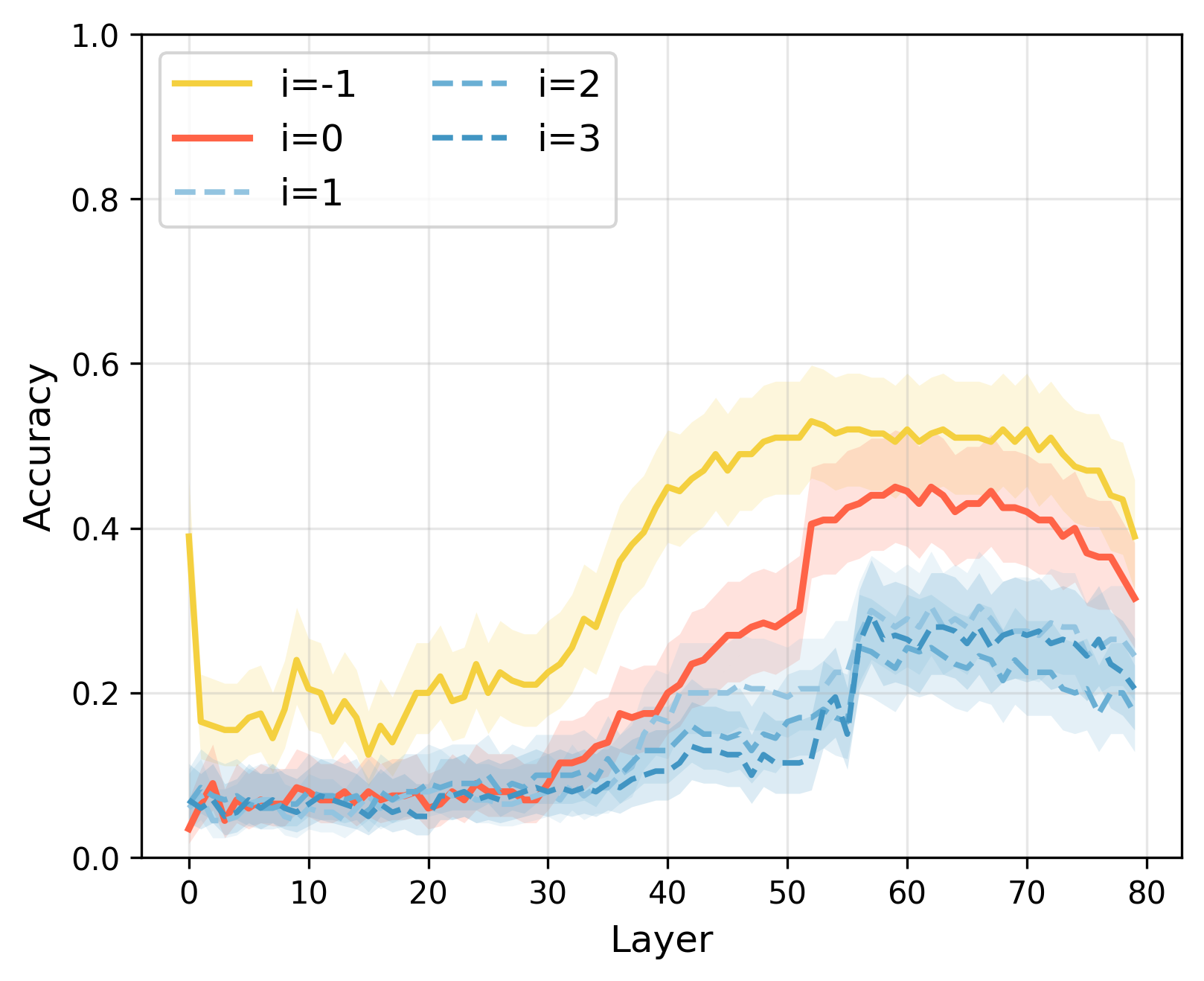}
    \end{subfigure}
    \centering
    \begin{subfigure}{0.325\linewidth}
        \centering
        \subcaption[]{Qwen3-32B (Rhyme Acc.)}
        \includegraphics[width=\linewidth]{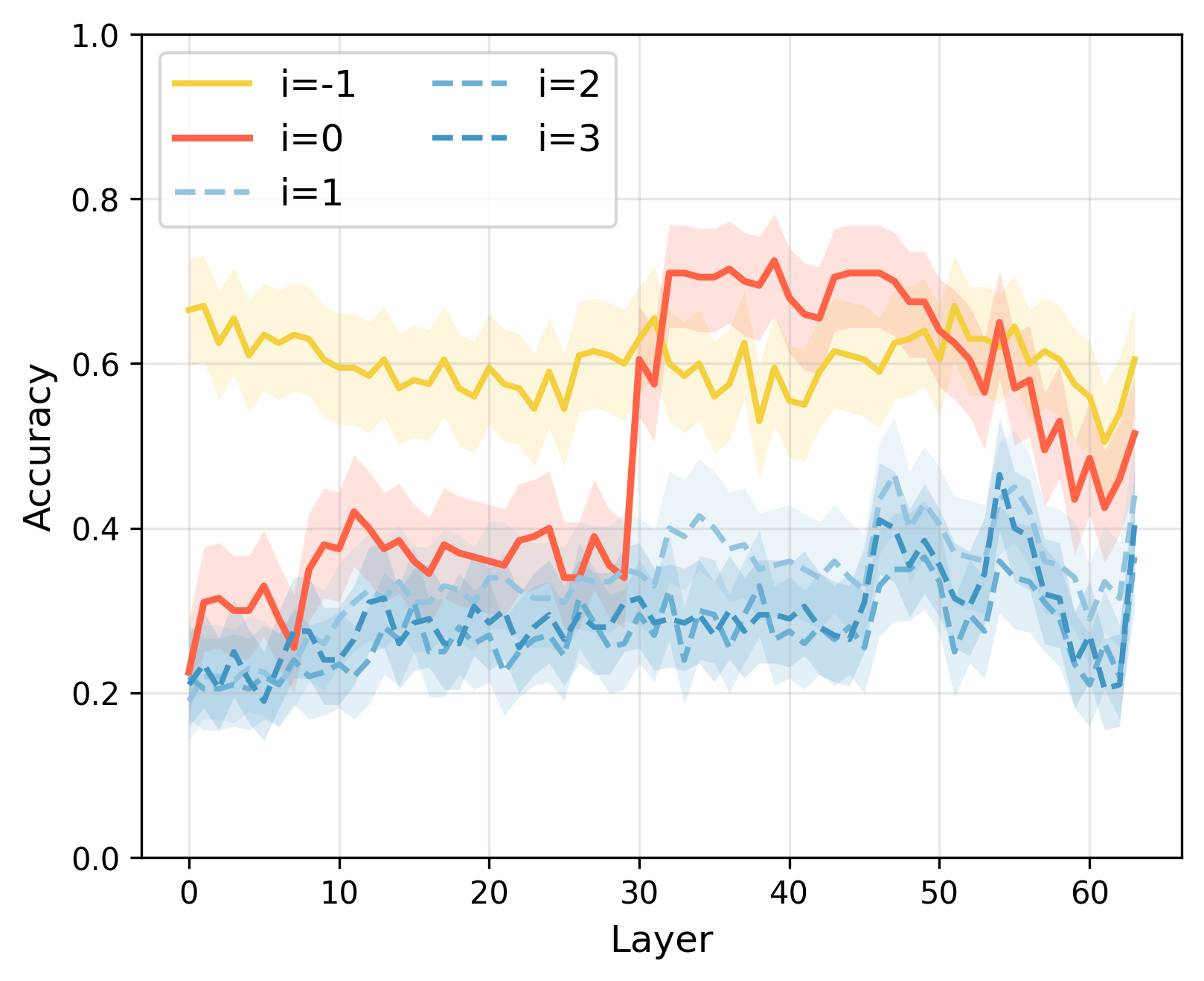}
    \end{subfigure}
    \begin{subfigure}{0.325\linewidth}
        \centering
        \subcaption[]{Gemma-3-27B (Rhyme Acc.)}
        \includegraphics[width=\linewidth]{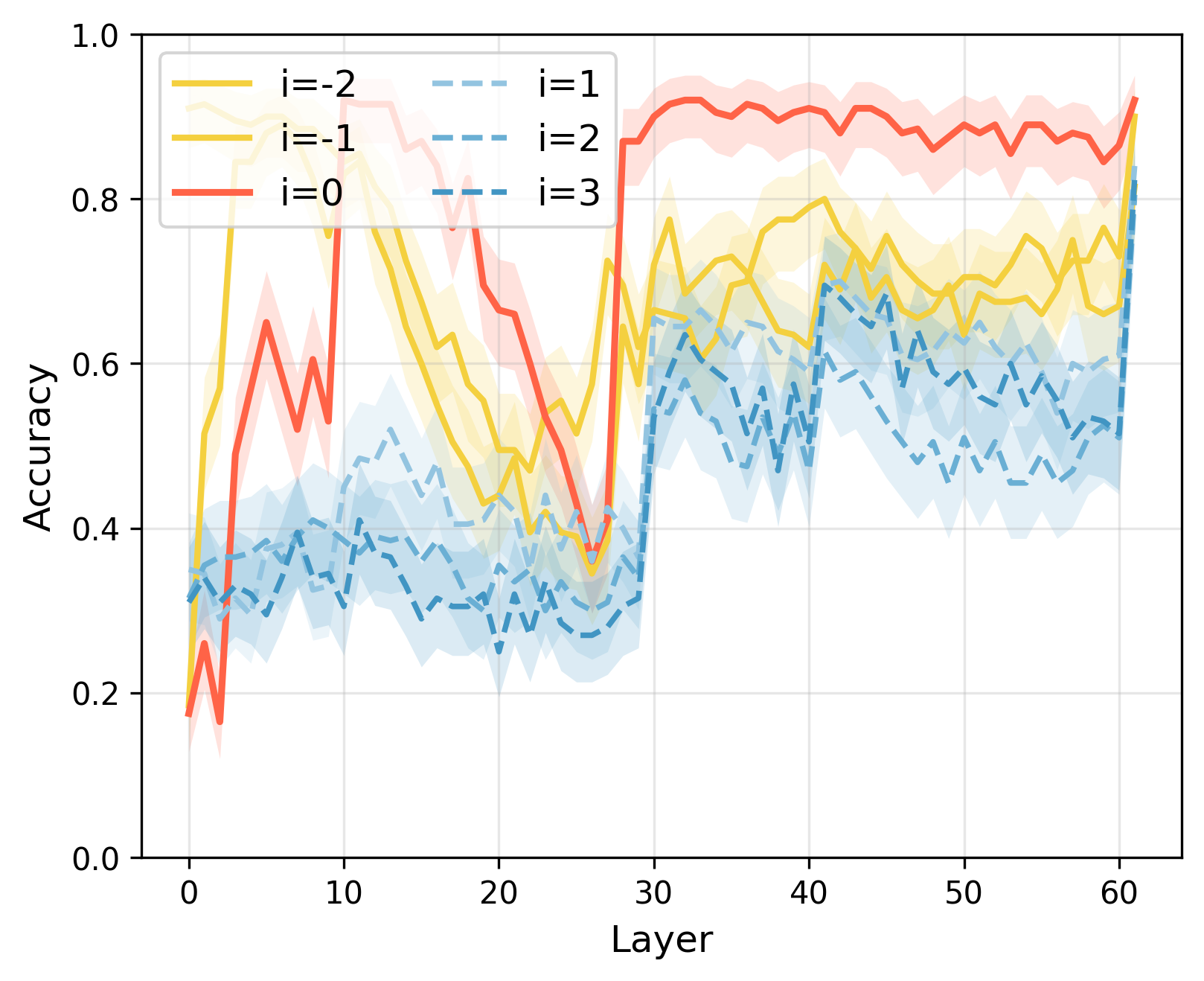}
    \end{subfigure}
    \begin{subfigure}{0.325\linewidth}
        \centering
        \subcaption[]{Llama-3.1-70B (Rhyme Acc.)}
        \includegraphics[width=\linewidth]{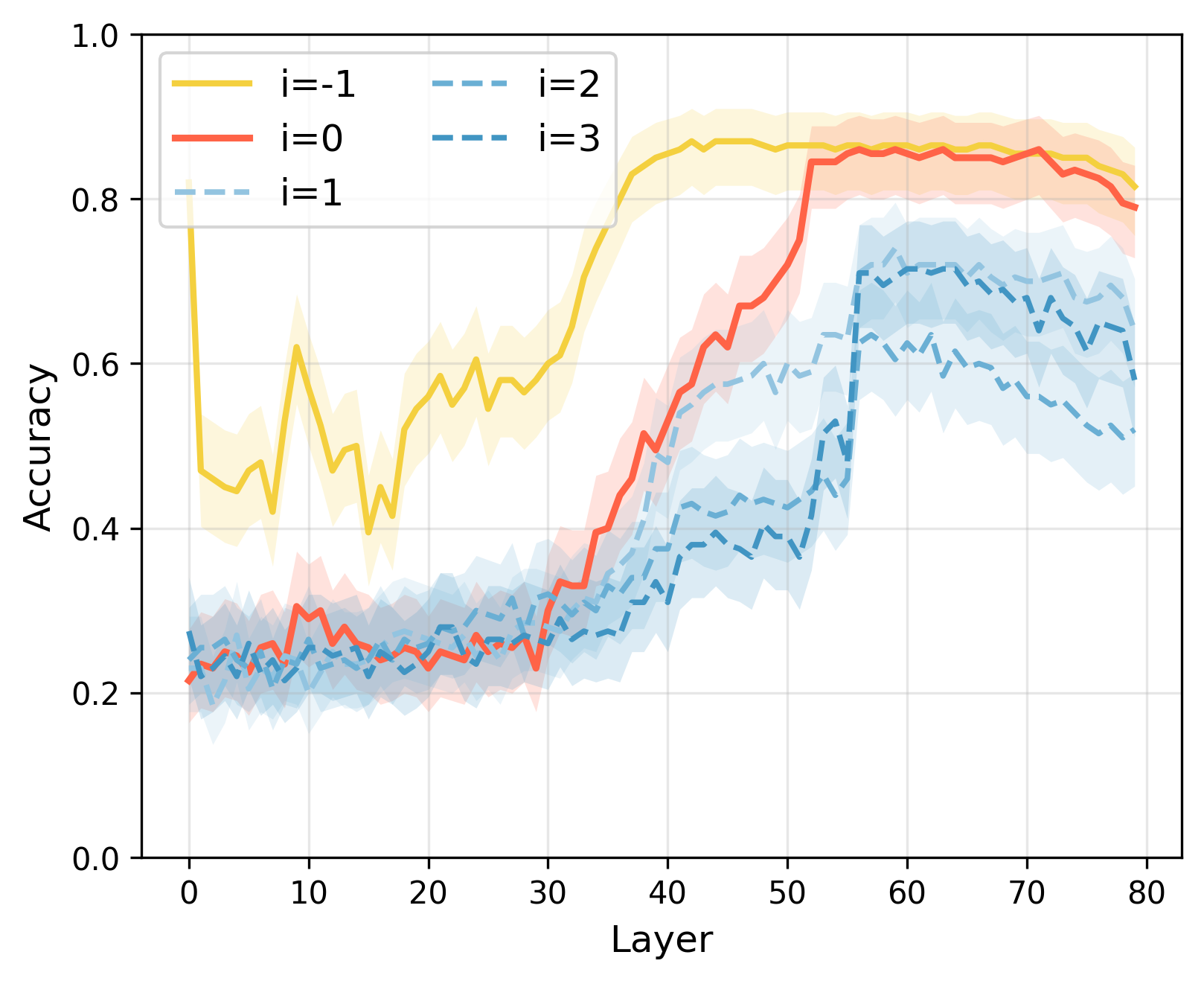}
    \end{subfigure}
    \caption{Top-5 and rhyme accuracy of linear probes trained to predict $r_2$ from hidden states at various layers and positions. Shaded bands are Wilson 95\% CIs computed from $N=200$ validation items. Probes at the last word ($i\leq-1$) and newline ($i=0$) positions substantially outperform probes at subsequent generated positions ($i>0$); the bands at the peak layers do not overlap with $i\geq1$ bands in any of the three models, indicating that rhyme-relevant information is selectively concentrated at these structural positions.}
    \label{fig:poem-results}
\end{figure*}

First, probe accuracies for predicting $r_2$ are substantially higher than for general-text continuation. Although the generated $r_2$ is on average 8 tokens past the newline (the last context token), the $i=0$ couplet probe far outperforms the corresponding $k=8$ Pile probe. This suggests that models selectively construct planning-compatible representations in response to the structural demands of rhyming couplet generation.

More importantly, the signal is concentrated at the last word and newline positions in specific layers (Figure~\ref{fig:poem-results}): the $i\leq0$ probes outperform the $i>0$ probes by a wide margin, in contrast to the monotonic decay with look-ahead distance seen on general text.

Probes at the last word peak early and decay later, consistent with the lexical identity of $r_1$ being encoded from the earliest layers. The newline probes are more striking: they also perform well, hinting at latent planning at this position. Whereas the $i>0$ probes follow a similar shape throughout, the $i\leq0$ probes trace a qualitatively different curve across layers---evidence that the newline and last word undergo specialized computations rather than passively accumulating context.

\begin{figure*}[t]
    \centering
    \begin{subfigure}{0.49\linewidth}
        \centering
        \includegraphics[width=\linewidth]{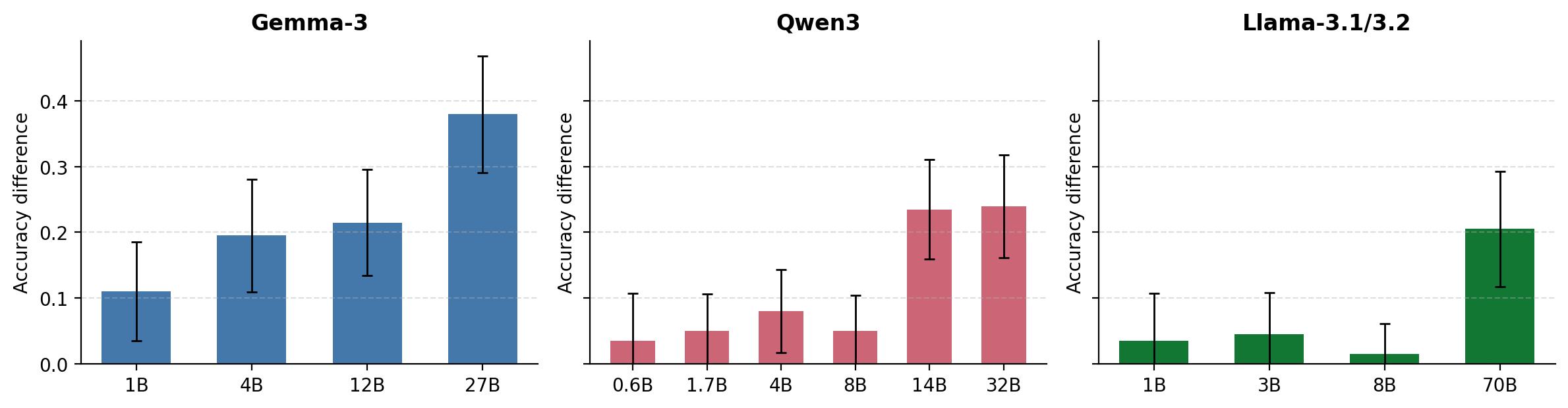}
        \subcaption[]{Top-5 Accuracy Difference}
    \end{subfigure}
    \begin{subfigure}{0.49\linewidth}
        \centering
        \includegraphics[width=\linewidth]{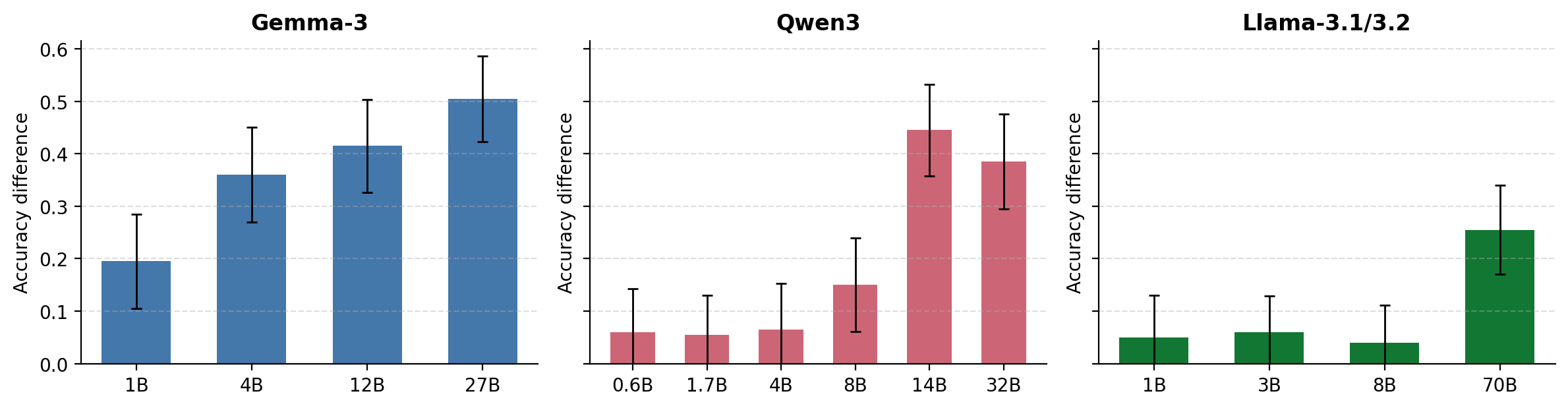}
        \subcaption[]{Rhyme Accuracy Difference}
    \end{subfigure}
    \caption{Maximum accuracy gap across layers between probes at the newline ($i=0$) and the first generated position ($i=1$), plotted against model size. Black error bars are 95\% CIs at the chosen peak layer (paired-difference Wald approximation; per-sample correctness was not stored, so the interval is conservative relative to the true paired CI). For Qwen3 sizes 0.6B--8B and Llama-3 sizes 1B--8B, the CI on the gap includes zero, while every Gemma-3 size and the largest Qwen/Llama models have CIs strictly above zero, supporting the emergent-property interpretation.}
    \label{fig:poem-scaling}
\end{figure*}

Replicating this experiment on smaller Qwen, Gemma, and Llama models reveals that planning-compatible representations at the newline emerge with scale. For each model we measure the largest accuracy gap across layers between probes at the newline and probes at the first generated position; this gap grows with scale (Figure~\ref{fig:poem-scaling}). Within Qwen3 and Llama-3, the gap CIs of the smaller models (Qwen3 0.6B--8B; Llama-3 1B--8B) all overlap zero, whereas the largest model in each family has a clearly nonzero gap. Gemma-3 shows a positive gap at every scale and the cleanest monotonic trend, rising from 0.11 at 1B to 0.38 at 27B. This pattern is consistent with planning-compatible representations being an emergent property of larger models.

\subsection{Discussion and limitations of probing}
\label{sec:probe-discussion}

The probing results reveal substantial cross-model and cross-scale diversity in planning-compatible representations at both the last word and newline positions. However, several alternative explanations warrant consideration.

First, the elevated newline probe accuracy could reflect passive attention accumulation rather than active planning: if models attend strongly from the newline to the rhyme word, the newline hidden state will encode rhyme-relevant information without any downstream computation reading it out during generation. Second, rhyme accuracy is an imperfect metric because the CMU Pronouncing Dictionary does not cover every valid rhyme, potentially underestimating probe performance unevenly across rhyme families. Finally, although we prompted Claude for diverse subjects and rhyme schemes, the resulting dataset may carry nontrivial distributional biases that could inflate probe accuracy in ways difficult to detect.

These considerations highlight two fundamental limitations of probing as a method. First, high probe accuracy establishes that rhyme-relevant information is linearly decodable at a position, but cannot establish that this information causally drives generation. Second, probe accuracy is sensitive to the distribution of the training data, meaning results could partially reflect dataset artifacts rather than genuine planning representations.

Activation patching addresses both limitations directly: by intervening on specific hidden states during generation and measuring the causal effect on output, it neither requires a probe training distribution nor conflates passive encoding with active deployment. We turn to activation patching in the next section to resolve these ambiguities.

\section{Patching for Causally Active Planning Sites}
\label{sec:patching}

\begin{figure}[t]
    \centering
    \includegraphics[width=\linewidth]{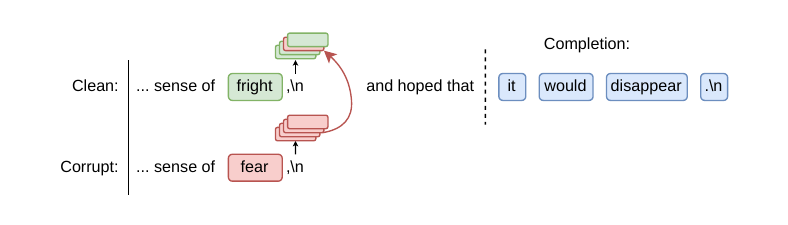}
    \caption{Activation patching: the hidden state $\mathbf{h}_{\ell,i}$ from a corrupted run is substituted into the clean run's forward pass at position $i$ and layer $\ell$. A successful patch redirects generation toward the corrupted rhyme family, providing causal evidence that $(i,\ell)$ is a planning site.}
    \label{fig:patch-diagram}
\end{figure}

While probing reveals that future token information is encoded in hidden states, it does not establish that this information is causally used by the model. In this section, we further investigate latent planning in rhyming couplet generation by employing intervention methods to test for causal influence directly. Given a prompt $\mathbf{x}^{(c)}$ whose first line naturally leads to a clean rhyme family $\mathcal{R}^{(c)}$, we intervene on the model's activations at various positions and investigate whether the generation of the second line shifts toward a different corrupted rhyme family $\mathcal{R}^{(r)}$. A successful intervention on a specific layer $\ell$ and position $i$ provides causal evidence that information at $\mathbf{h}_{\ell, i}$ is causally involved in the model's latent planning downstream.

\begin{figure*}[t]
    \centering
    \begin{subfigure}{0.32\linewidth}
        \centering
        \includegraphics[width=\linewidth]{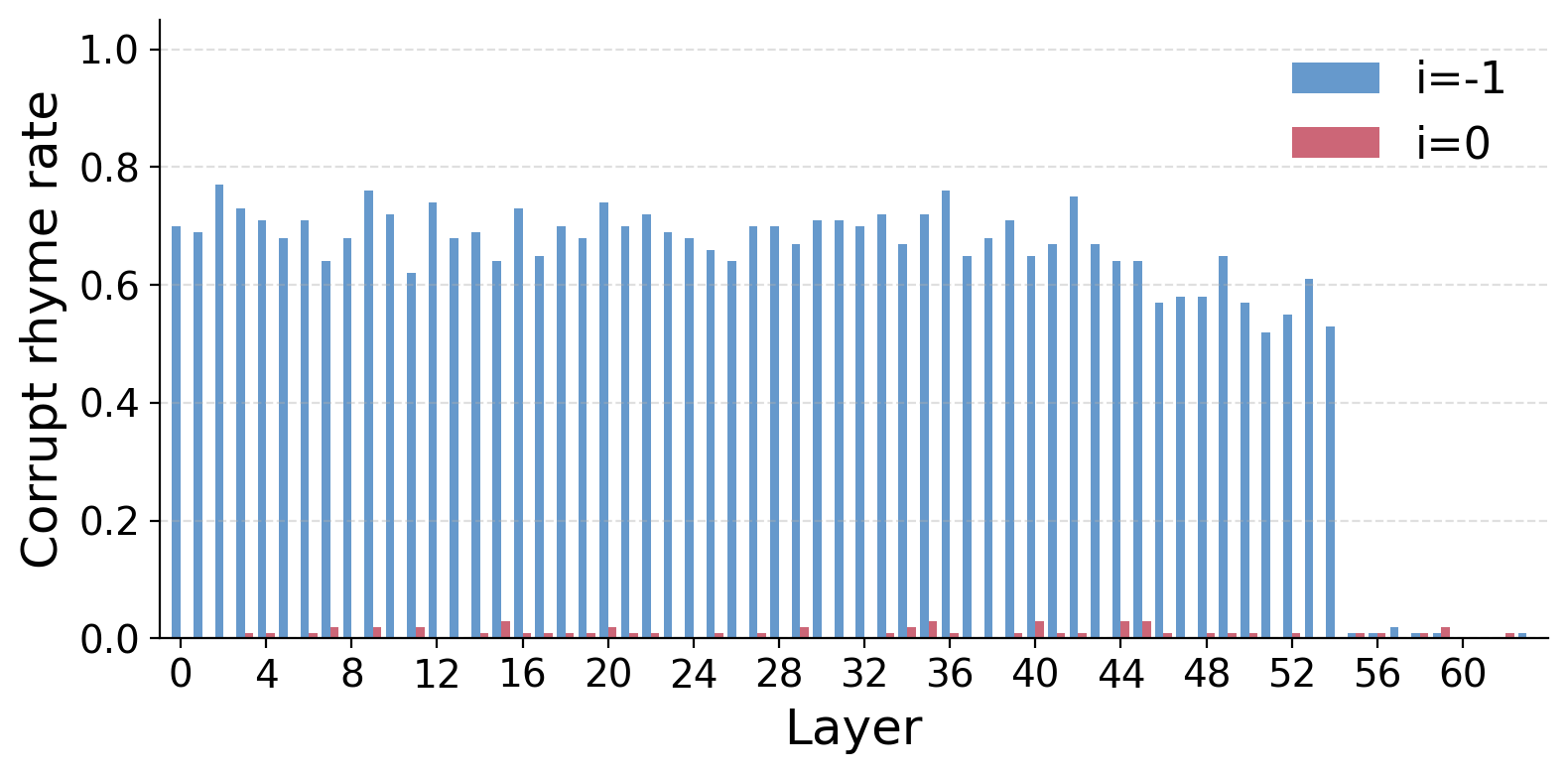}
        \caption{Qwen3-32B.}
        \label{fig:patch-qwen}
    \end{subfigure}
    \hfill
    \begin{subfigure}{0.32\linewidth}
        \centering
        \includegraphics[width=\linewidth]{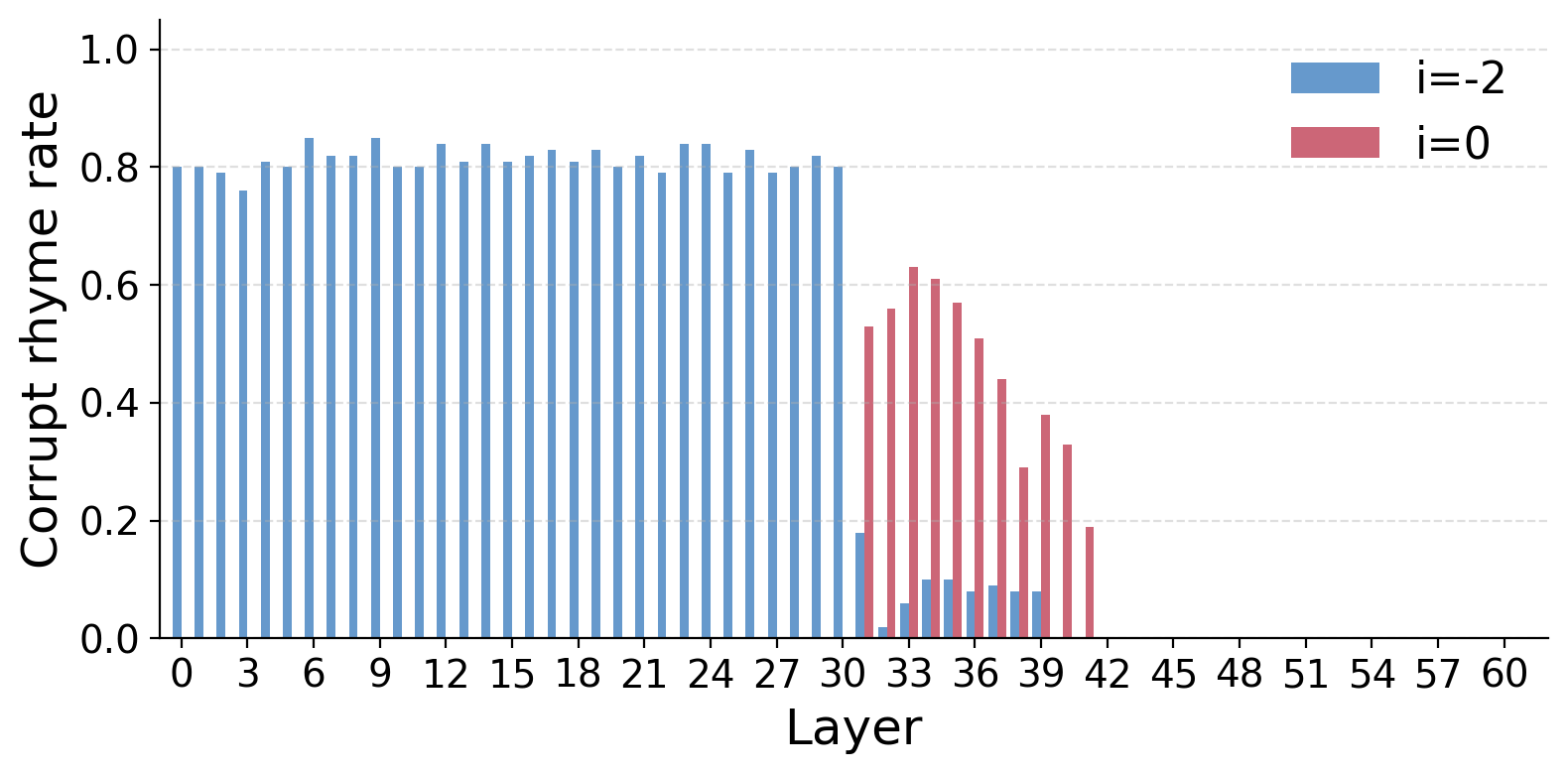}
        \caption{Gemma-3-27B.}
        \label{fig:patch-gemma}
    \end{subfigure}
    \hfill
    \begin{subfigure}{0.32\linewidth}
        \centering
        \includegraphics[width=\linewidth]{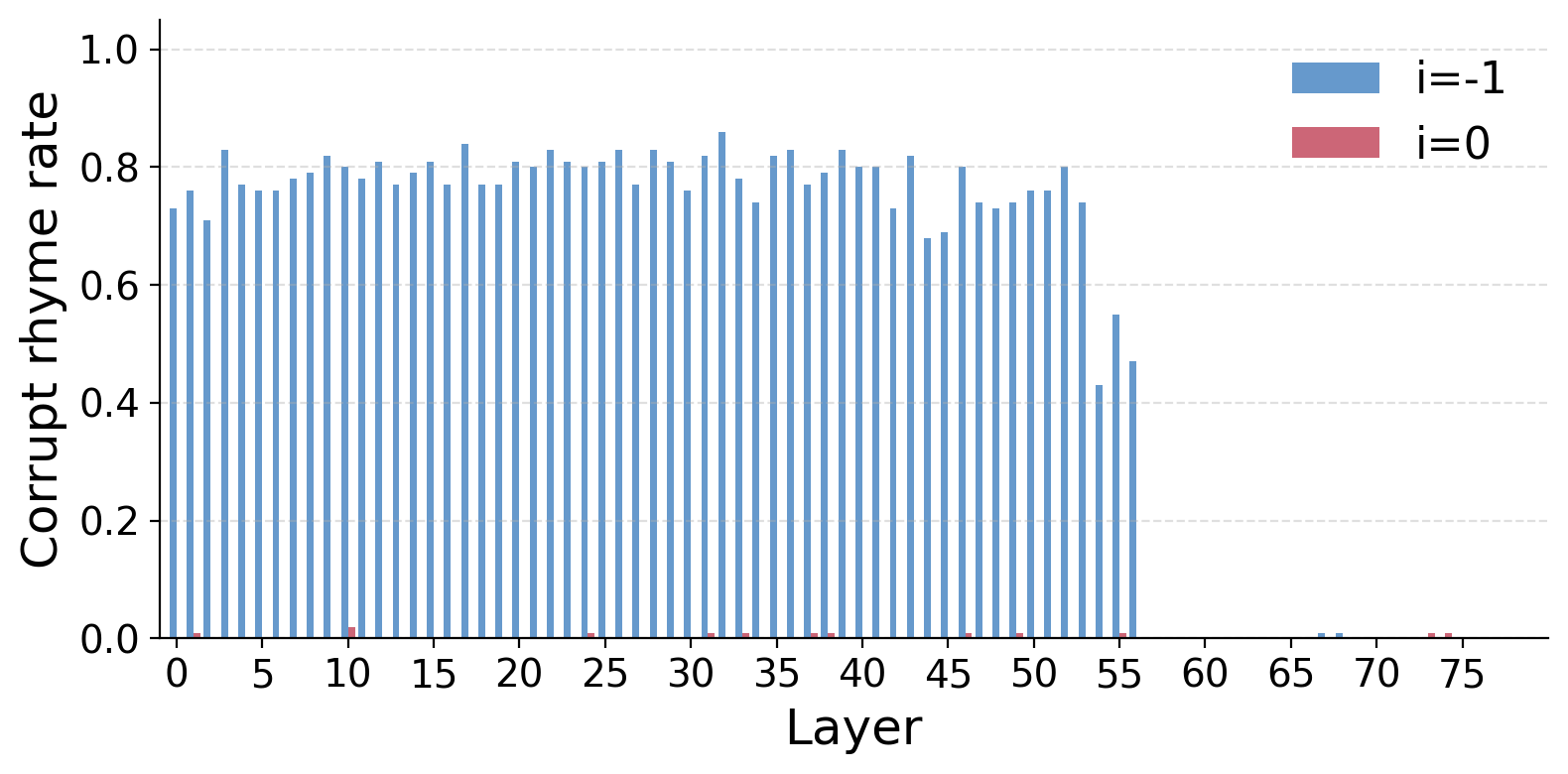}
        \caption{Llama-3.1-70B.}
        \label{fig:patch-llama}
    \end{subfigure}
    \caption{Per-layer activation patching at the last word token and newline ($i=0$) for the largest model in each family. Full results for all model sizes are in Appendix~\ref{sec:additional-patching-details}.}
    \label{fig:patch-results}
\end{figure*}

We apply activation patching at two positions: the last word token (position $i=-1$ in Qwen3 and Llama-3, $i=-2$ in Gemma-3 due to its tokenization of the line-ending comma; see Appendix~\ref{sec:model-config}) and the newline token ($i=0$), sweeping across all layers individually. For each layer, we draw $N=20$ stochastic samples per prompt pair; the main per-layer figures (Figure~\ref{fig:patch-results}) average over 5 prompt pairs ($N{=}100$). Because the same word pairs are reused across layers and positions, the prompt pair is the natural unit of independence; we report 95\% cluster bootstrap intervals (10{,}000 pair-level resamples) on every patching rate, and use a joint cluster bootstrap with shared pair indices when comparing two patching conditions on the same pairs.

In Gemma-3-27B (Figure~\ref{fig:patch-gemma}), last-word patching is highly effective in early layers but drops sharply around layer 30, while newline patching rises simultaneously to a peak corrupt rhyme rate of 0.63 [95\% CI 0.48, 0.78] at layer 33. We term this crossover the \textit{representational handoff}: the causal locus migrates from the last word token to the newline, which becomes the primary read-out site for the phonological constraint. The wide CI reflects pair-to-pair variability (per-pair rates at layer 33 span 0.40--0.90), but the qualitative migration is consistent across pairs. By contrast, Qwen3-32B (Figure~\ref{fig:patch-qwen}) and Llama-3.1-70B (Figure~\ref{fig:patch-llama}) show uniformly high last-word patching and near-zero newline patching across every layer, with non-overlapping CIs separating the two positions. These models condition on the last word token throughout generation. Swapping activation patching for a steering vector intervention~\citep{turner2023activation,maar2026whatsplanmetricsimplicit} gives the same three-model picture: Gemma-3-27B hands off to the newline around layer 30, and Qwen3-32B and Llama-3.1-70B remain effective at the last word but flat at the newline. Steering needs far more data than patching to reach this conclusion (Appendix~\ref{sec:steering}). Per-layer patching results for every model size are in Appendix~\ref{sec:additional-patching-details}.

\subsection{Localizing the Handoff to a Sparse Set of Attention Heads}
\label{sec:head-patching}

\begin{figure*}[!t]
    \centering
    \begin{subfigure}{0.49\linewidth}
        \centering
        \includegraphics[width=\linewidth]{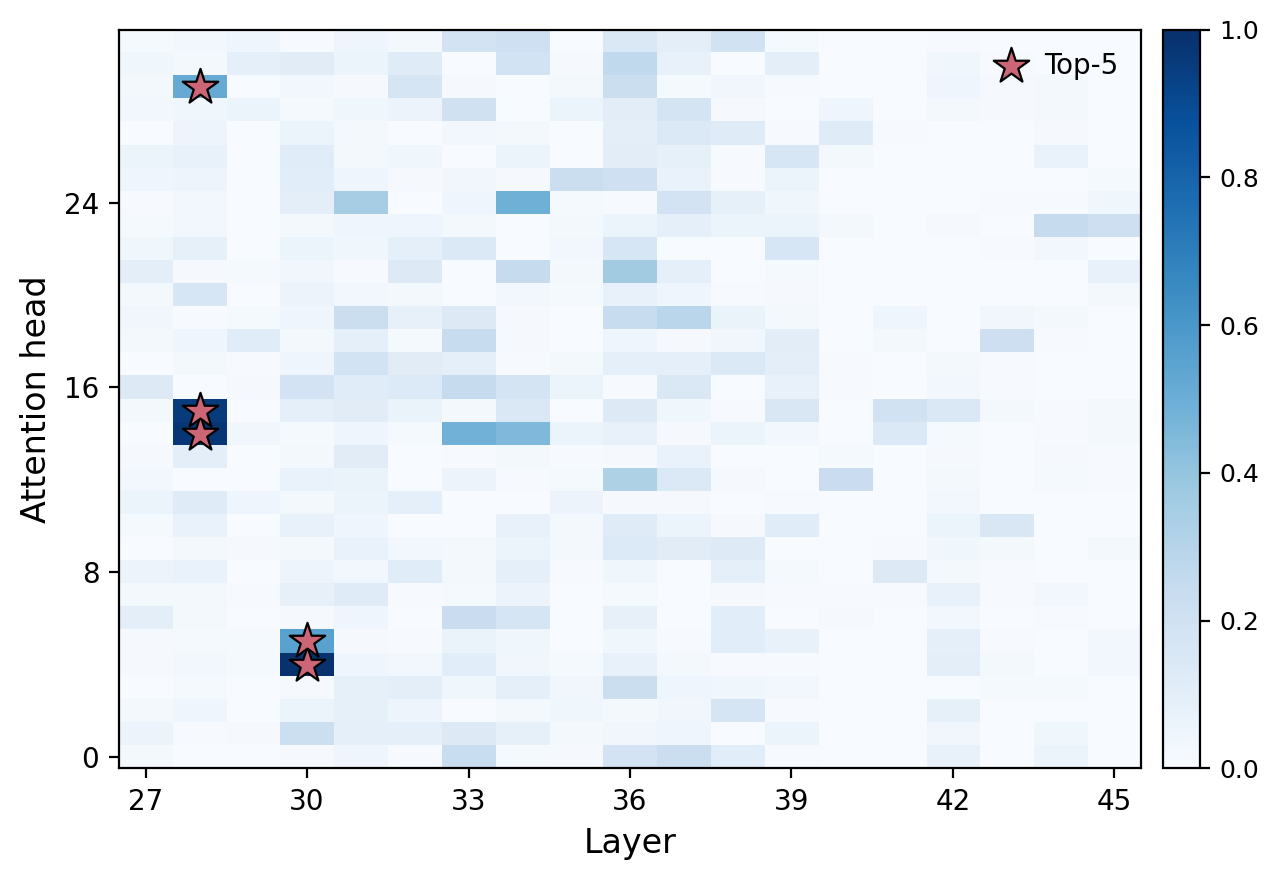}
        \subcaption{Attention weight, $i{=}0 \to i{=}{-}2$.}
        \label{fig:topk-heads-a}
    \end{subfigure}
    \hfill
    \begin{subfigure}{0.49\linewidth}
        \centering
        \includegraphics[width=\linewidth]{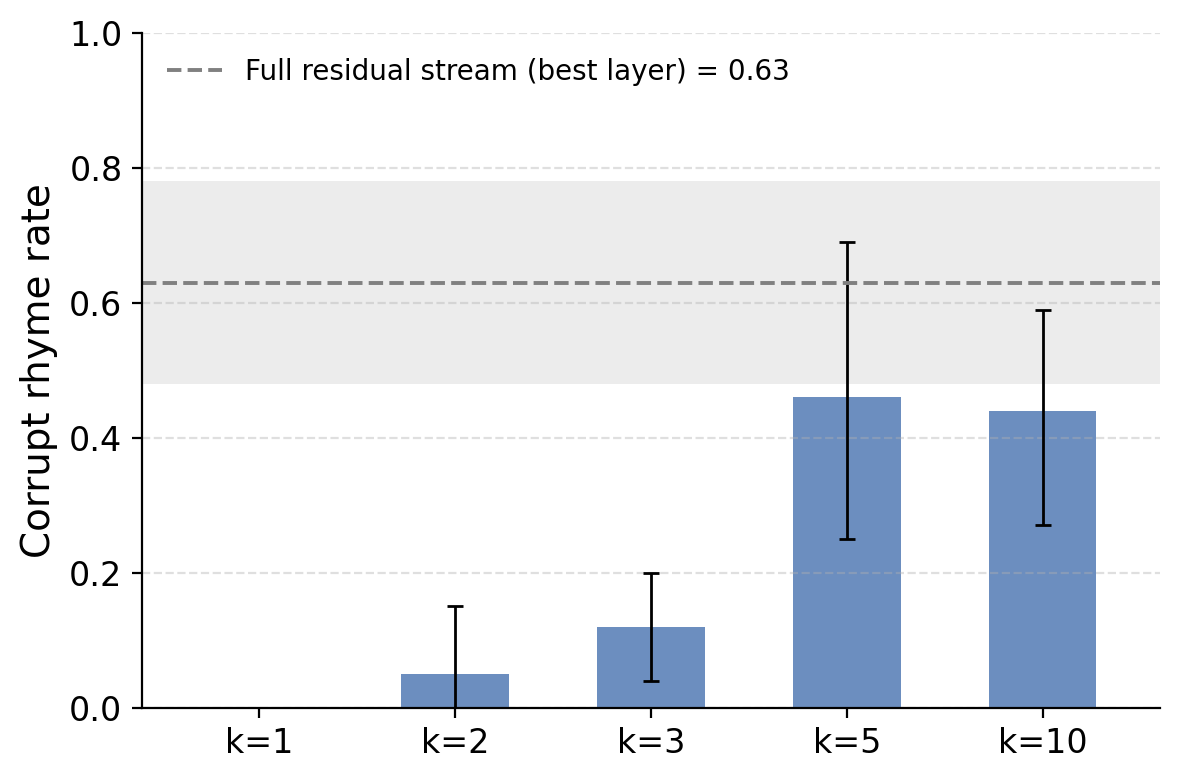}
        \subcaption{Top-$k$ head patching at the newline.}
        \label{fig:topk-heads-b}
    \end{subfigure}
    \caption{Localizing the planning site handoff in Gemma-3-27B to a sparse set of attention heads. \textbf{(a)} Attention weight from the newline token ($i=0$) to the last word token ($i=-2$) across heads in layers 27--45. Red stars mark the top-5 heads by attention weight. \textbf{(b)} Corrupt rhyme rate when patching the top-$k$ highest-attending heads simultaneously at the newline (5 prompt pairs $\times\ N{=}20$). Black error bars are cluster-bootstrap 95\% CIs over pairs. The dashed line and shaded gray band show the full-residual-stream peak at the best single layer (5-pair estimate $0.63$ [0.48, 0.78]).}
    \label{fig:topk-heads}
\end{figure*}

Having established that the planning site forms at the newline token in Gemma-3-27B, we ask whether the information routing handoff can be attributed to a small, identifiable set of attention heads. Single-head patching at the newline (replacing one head's output at a time within the planning layer range, layers 27--45) produced no measurable signal for any individual head, suggesting the representation is not localized to a single circuit element. We use attention weights as a proxy for which heads are most likely to route rhyme information from the last word token to the newline.

Specifically, we extract the attention weight from the newline token ($i=0$) to the last word token ($i=-2$) for each head in layers 27--45 of both clean and corrupt forward passes, and rank heads by this weight. Figure~\ref{fig:topk-heads}a shows the resulting heatmap. Attention to $i=-2$ is highly concentrated in three heads that attend almost exclusively to the last word token from the newline---layer 30 head 4 (weight ${\approx}0.99$), layer 28 head 14 (${\approx}0.97$), and layer 28 head 15 (${\approx}0.95$). A second cluster of heads in layers 28--36 shows moderate attention weights (0.35--0.55).

We patch the top-$k$ heads simultaneously, replacing their outputs at the newline with what they would have produced on the corrupt forward pass, and measure the resulting corrupt rhyme rate. To interpret these rates we compare against the strongest sub-component intervention available at the newline: replacing the entire residual stream at $i{=}0$ with corrupt's residual at the best single layer. Because the residual at $i{=}0$ is the sum of every attention head and MLP contribution up to that layer, this overwrites all of them at once and provides a strong reference for what any subset of components feeding $i{=}0$ recovers in practice (a subset patch can in principle exceed it when the full residual also injects conflicting context). We call this rate the \textit{full-residual reference} (Section~\ref{sec:patching}); for Gemma-3-27B it is 63\%.

\textit{Simple top-$k$ patching.} Figure~\ref{fig:topk-heads}b shows that $k=1,2,3$ yield near-zero corrupt rhyme rates. At $k=5$ the rate jumps to 46\%, which is 73\% of the full-residual reference, and plateaus through $k=10$. This means that patching just five attention heads at the newline reproduces about three-quarters of the rhyme-shifting effect of overwriting the entire residual at the best layer.

\begin{figure}[H]
    \centering
    \includegraphics[width=\linewidth]{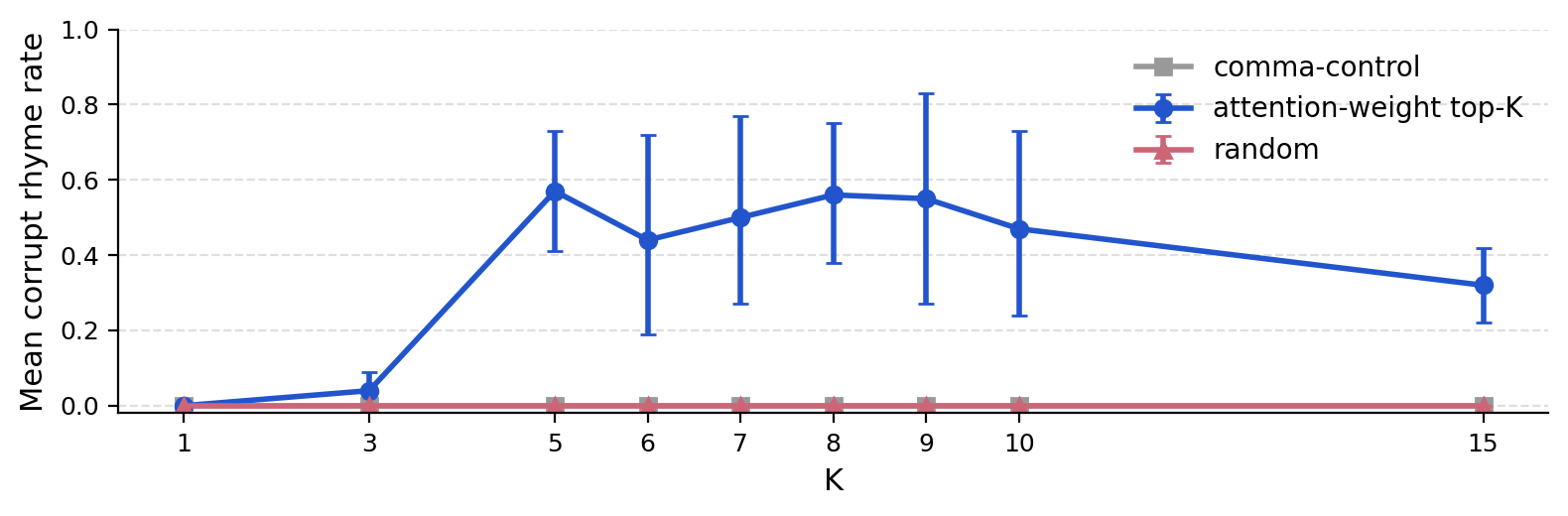}
    \caption{Two-stage path patching K-sweep on Gemma-3-27B. Attention-weight top-$k$ peaks at $k=5$ (57\%, 90\% of the full-residual reference) and declines at $k=10, 15$. Comma-control and random head sets stay at zero. Error bars are 95\% cluster-bootstrap CIs over prompt pairs.}
    \label{fig:path-patch-curve}
\end{figure}

\textit{Two-stage path patching.} To isolate the specific path $i{=}-2 \to \text{head} \to i{=}0 \to \text{output}$, we apply two-stage path patching~\citep{goldowsky2023localizing,wang2022ioi}. Stage~1: run the clean prompt with only the residual at $i{=}-2$ replaced by corrupt's, and cache each candidate head's output at $i{=}0$. Stage~2: forward the unmodified clean prompt and substitute those cached outputs at $i{=}0$ for the selected heads. Under this stricter intervention (Figure~\ref{fig:path-patch-curve}), the five heads recover a corrupt rhyme rate of 57\% at $k=5$, or 90\% of the full-residual reference. This means that nearly all of the rhyme-routing capacity at the newline is concentrated in these five heads. The rate stays in the 44--57\% range over $k=5{-}9$, with $k=5$ the highest individual point, and declines at $k=10$ and $k=15$ (47\% and 32\%). Random and comma-control head sets remain at zero across all $k$.

The top five heads are (layer 30, head 4), (layer 28, head 14), (layer 28, head 15), (layer 30, head 5), and (layer 28, head 29). The analogous top-$k$ MLP patches at $i{=}0$ yield zero corrupt rhyme rate at every $k$ (Wilson 95\% upper bound $\leq 0.04$, $N{=}100$), confirming that the handoff is mediated by attention rather than feed-forward computation.

\section{Conclusion}

We introduced \textit{planning site formation} and studied it across three open-source model families using linear probing and activation patching. Across our experiments, encoding and use turn out to be dissociable. Probes detect rhyme-relevant information at the newline in many models and many scales, yet only Gemma-3-27B treats that information as causal during generation. Every other model we tested conditions on the rhyme word throughout, despite scale-dependent probe signal at the newline. Probe signal is not by itself evidence of a planning site.

When the handoff does occur it is implemented by a sparse, identifiable mechanism. In Gemma-3-27B we trace it to five attention heads in layers 28 and 30: the $k$-sweep peaks at exactly that set, declining as more heads are added, while random and comma-control head sets stay at zero across all $k$. Planning site formation, when it appears, is a structured computational phenomenon rather than a diffuse property of many components.

Our activation patching approach requires no extraneous model training and far less data than previous steering vector or transcoder-based approaches, making it scalable to large models across many architectures. The results, however, also surface important open questions and limitations that future work should address.

First, our analysis is limited to three model families on a single structured generation task. Extending the methodology to prose generation, code completion, and multi-step reasoning tasks would test whether the representational handoff is a general planning primitive or a narrow phonological phenomenon. Second, while two-stage path patching localizes the handoff to a five-head set, the wide upper bound on the recovered fraction (CI extending past 1.0) reflects the small number of prompt pairs. A larger, more diverse couplet set would tighten this estimate and reveal whether the head set varies by rhyme family. Third, the absence of the handoff in all Qwen3 and Llama-3 models despite strong probe signal raises a question about what distinguishes Gemma-3-27B architecturally or by training. Finally, causal scrubbing~\citep{chan2022causal} or activation steering experiments targeting the planning heads would help distinguish whether the planning representation at the newline is genuinely read during generation or exerts influence only when artificially inserted via patching.

\bibliography{references}
\bibliographystyle{icml2026}

\appendix

\section{Model Configurations}
\label{sec:model-config}
For reference, we report details of the architectural differences between models.

\begin{table}[h]
\centering
\begin{tabular}{lccc}
\toprule
 & Qwen3-32B & Gemma-3-27B  & Llama-3.1-70B \\
\midrule
$|\mathcal V|$ & 151,936 & 262,208 & 128,256 \\
$d$            & 5,120   & 5,376   & 8,192  \\
$L$            & 64      & 62      & 80     \\
\bottomrule
\end{tabular}
\caption{Model architecture summary.}
\label{tab:models}
\end{table}

We also note important tokenization differences. Qwen and Llama treat \texttt{,\textbackslash n} as a single token, while Gemma treats
\texttt{,} and \texttt{\textbackslash n} as separate tokens.
This places the last word $r_1$ at position $-1$ in Qwen and Llama and position $-2$ in Gemma.

\begin{table}[h]
\centering
\small
\begin{tabular}{ccccc}
\toprule
 & $-2$ & $-1$ & $0$ & $1$ \\
\midrule
Qwen3   & \texttt{of}     & \texttt{fright} & \texttt{,\textbackslash n} & \texttt{and} \\
Llama-3 & \texttt{of}     & \texttt{fright} & \texttt{,\textbackslash n} & \texttt{and} \\
Gemma-3 & \texttt{fright} & \texttt{,}      & \texttt{\textbackslash n}  & \texttt{and} \\
\bottomrule
\end{tabular}
\caption{Tokenization of the line boundary across model families.}
\end{table}

\section{Steering Vectors}
\label{sec:steering}

\begin{figure}[H]
    \centering
    \includegraphics[width=\linewidth,page=2]{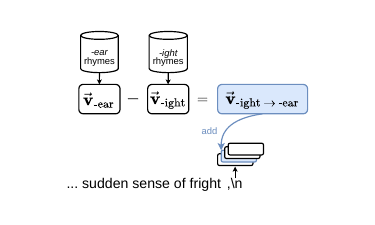}
    \caption{Steering vectors are mean-difference vectors between residual activations on prompts in two different rhyme schemes; adding $\alpha\mathbf{v}_{\ell,i}^{(s\to t)}$ at $(\ell, i)$ during generation should redirect the rhyme toward scheme $t$.}
    \label{fig:steer-diagram}
\end{figure}

We also replicate the main finding with a steering vector intervention~\citep{turner2023activation,maar2026whatsplanmetricsimplicit}. For each rhyme-scheme pair $(s, t)$ and each $(\ell, i)$, the steering vector $\mathbf{v}_{\ell,i}^{(s\to t)} = \bar{\mathbf{h}}_{\ell,i}^{(t)} - \bar{\mathbf{h}}_{\ell,i}^{(s)}$ is the mean difference of residual activations at $(\ell, i)$ between prompts ending in scheme $s$ and prompts ending in scheme $t$ (Figure~\ref{fig:steer-diagram}). At inference time we add $\alpha\mathbf{v}_{\ell,i}^{(s\to t)}$ at $(\ell, i)$ during generation on a held-out scheme-$s$ prompt and measure the fraction of completions that rhyme in scheme $t$. We use ten schemes, $\alpha = 1.5$, and the same CMU rhyme matcher as the patching evaluation.

Computing the vectors requires 10 schemes $\times$ 100 train prompts $=$ 1{,}000 hooked forward passes, and the evaluation sweep covers every $(\ell, i)$ across scheme pairs at 20 held-out prompts each. The patching runs in Section~\ref{sec:patching} use 5 prompt pairs $\times$ $N{=}20$ samples per cell. Same causal question, one to two orders of magnitude more data; we therefore treat steering as a cross-check rather than the primary method.

\begin{figure}[H]
    \centering
    \begin{subfigure}{\linewidth}
        \centering
        \subcaption[]{Qwen3-32B.}
        \includegraphics[width=\linewidth]{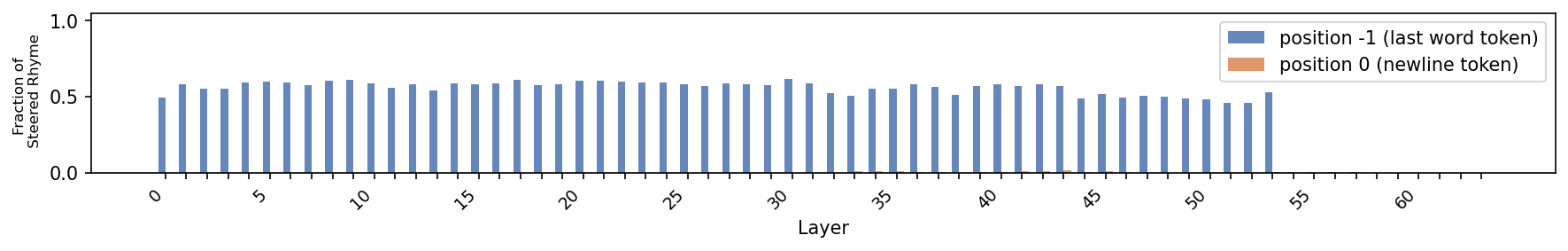}
    \end{subfigure}
    \begin{subfigure}{\linewidth}
        \centering
        \subcaption[]{Gemma-3-27B.}
        \includegraphics[width=\linewidth]{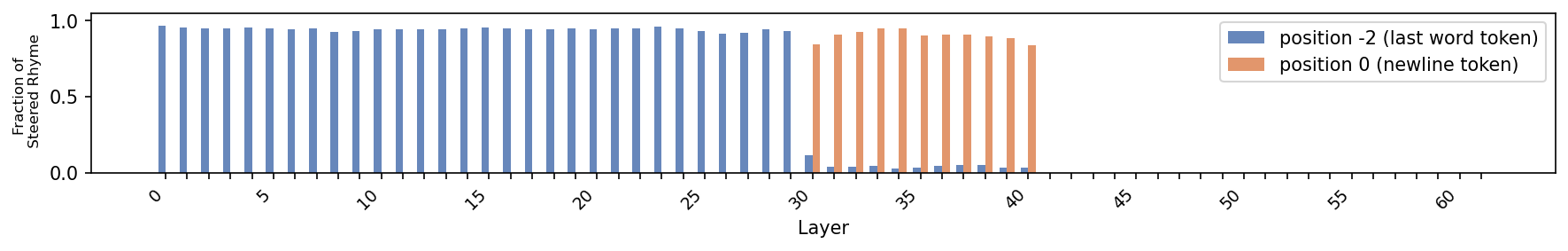}
    \end{subfigure}
    \begin{subfigure}{\linewidth}
        \centering
        \subcaption[]{Llama-3.1-70B.}
        \includegraphics[width=\linewidth]{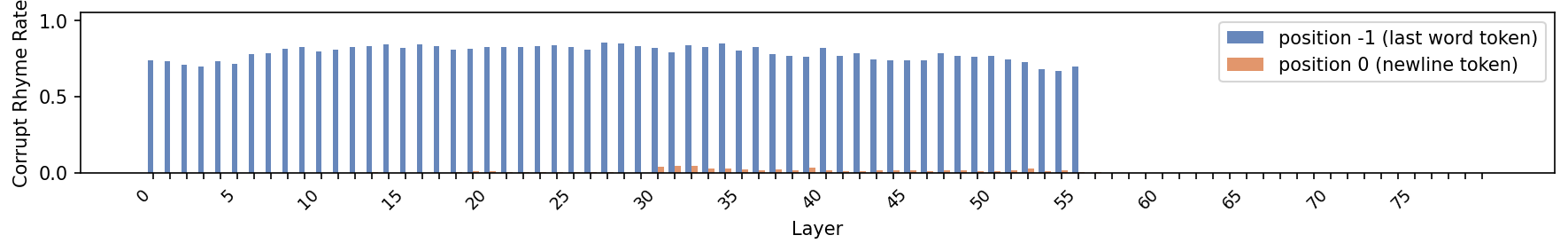}
    \end{subfigure}
    \caption{Steered rhyme fraction at the last word and newline across all layers, averaged over scheme pairs ($\alpha{=}1.5$, 100 train prompts per scheme). The picture mirrors Figure~\ref{fig:patch-results}: Gemma-3-27B shows the same handoff (last-word effect drops near layer 30 as the newline effect rises to $0.85$--$0.95$), while Qwen3-32B and Llama-3.1-70B show effective last-word steering at every swept layer with newline steering at noise.}
    \label{fig:steer-results}
\end{figure}

Figure~\ref{fig:steer-results} shows the steered rhyme fraction at the last word and newline across all layers for Qwen3-32B, Gemma-3-27B, and Llama-3.1-70B; the qualitative picture matches Figure~\ref{fig:patch-results}. In Gemma-3-27B, last-word steering is effective in early layers and drops sharply around layer 30 while newline steering rises simultaneously into the $0.85$--$0.95$ range across layers 30--40---the same handoff. Qwen3-32B and Llama-3.1-70B show effective last-word steering at every swept layer with newline steering at noise. The handoff and its absence both reappear under steering, and the rate estimates are tighter than the patching CIs because of the larger evaluation set.

\section{Additional Probing Results}
\label{sec:additional-probing-results}
For probing experiments, we also evaluated on top-1 accuracy. Note these are similar to the reported results, only scaled down.

\begin{figure}[H]
    \centering
    \begin{subfigure}{0.49\linewidth}
        \centering
        \subcaption[]{Qwen3-32B.}
        \includegraphics[width=\linewidth]{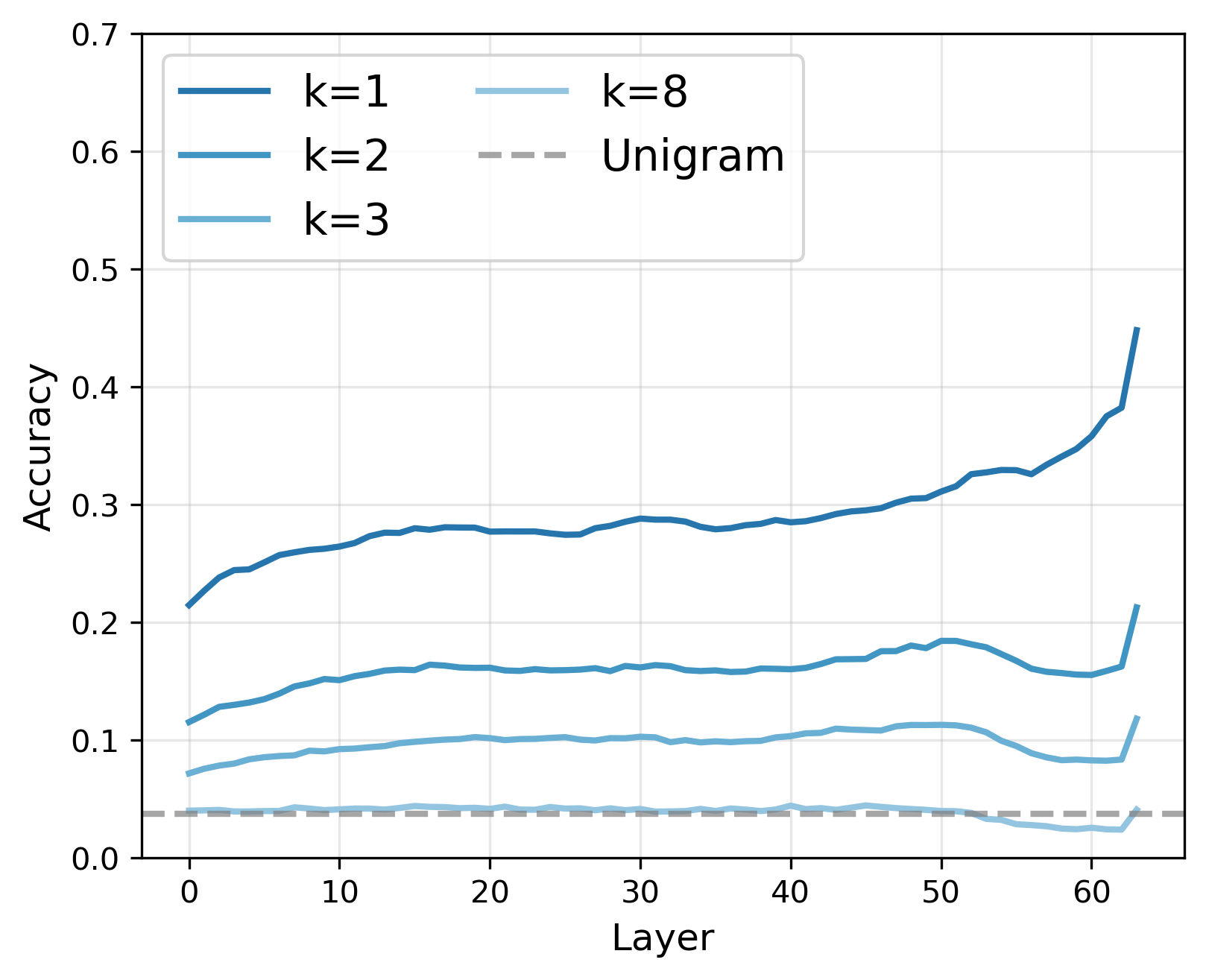}
    \end{subfigure}
    \begin{subfigure}{0.49\linewidth}
        \centering
        \subcaption[]{Gemma-3-27B.}
        \includegraphics[width=\linewidth]{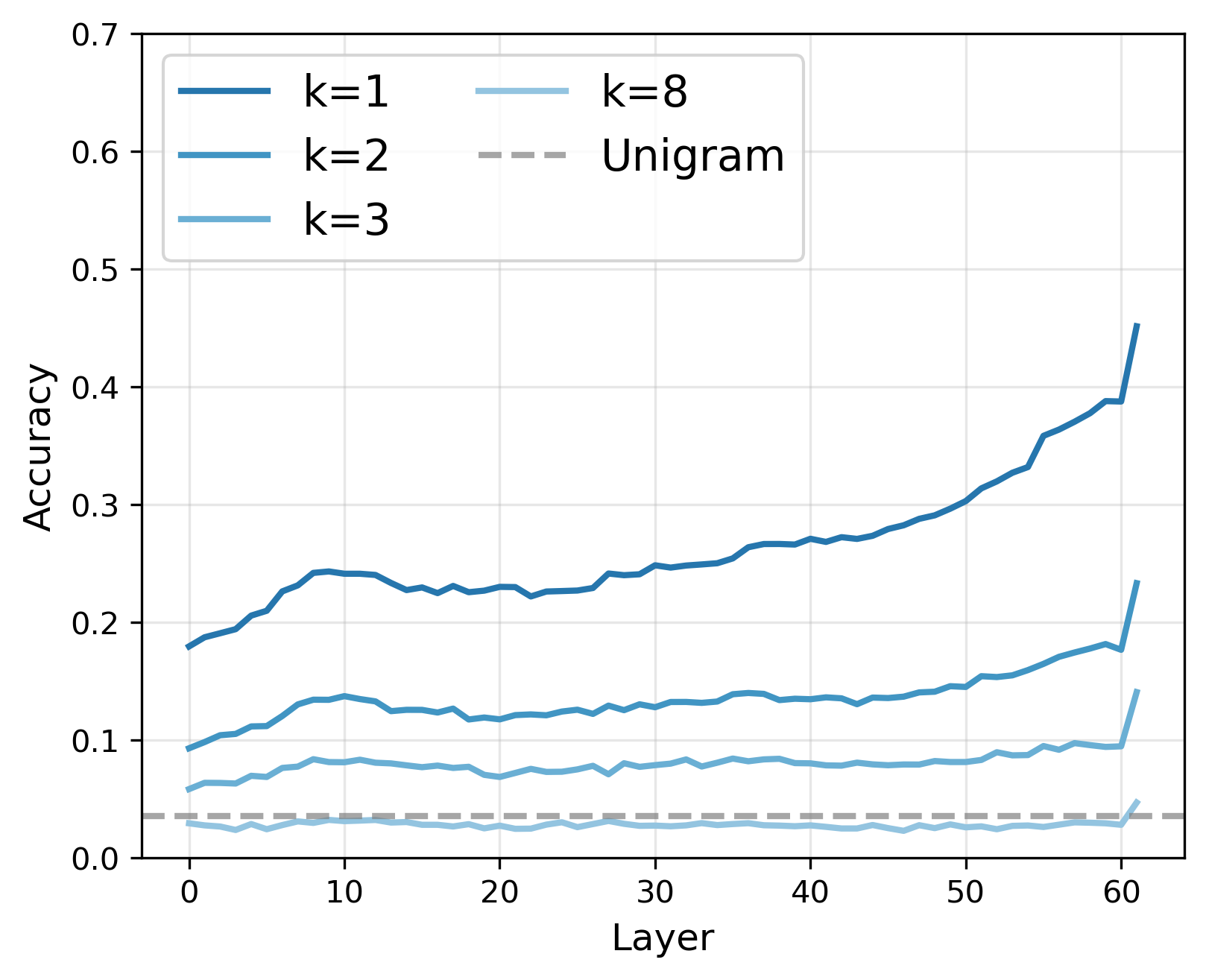}
    \end{subfigure}
    \caption{Top-1 probe accuracy predicting $k$ tokens ahead in general text (Pile). Mirrors the top-5 pattern: accuracy degrades monotonically with $k$ and falls to unigram baseline by $k=8$.}
    \label{fig:appendix-probe}
\end{figure}

\begin{figure}[H]
    \centering
    \begin{subfigure}{0.49\linewidth}
        \centering
        \subcaption[]{Qwen3-32B}
        \includegraphics[width=\linewidth]{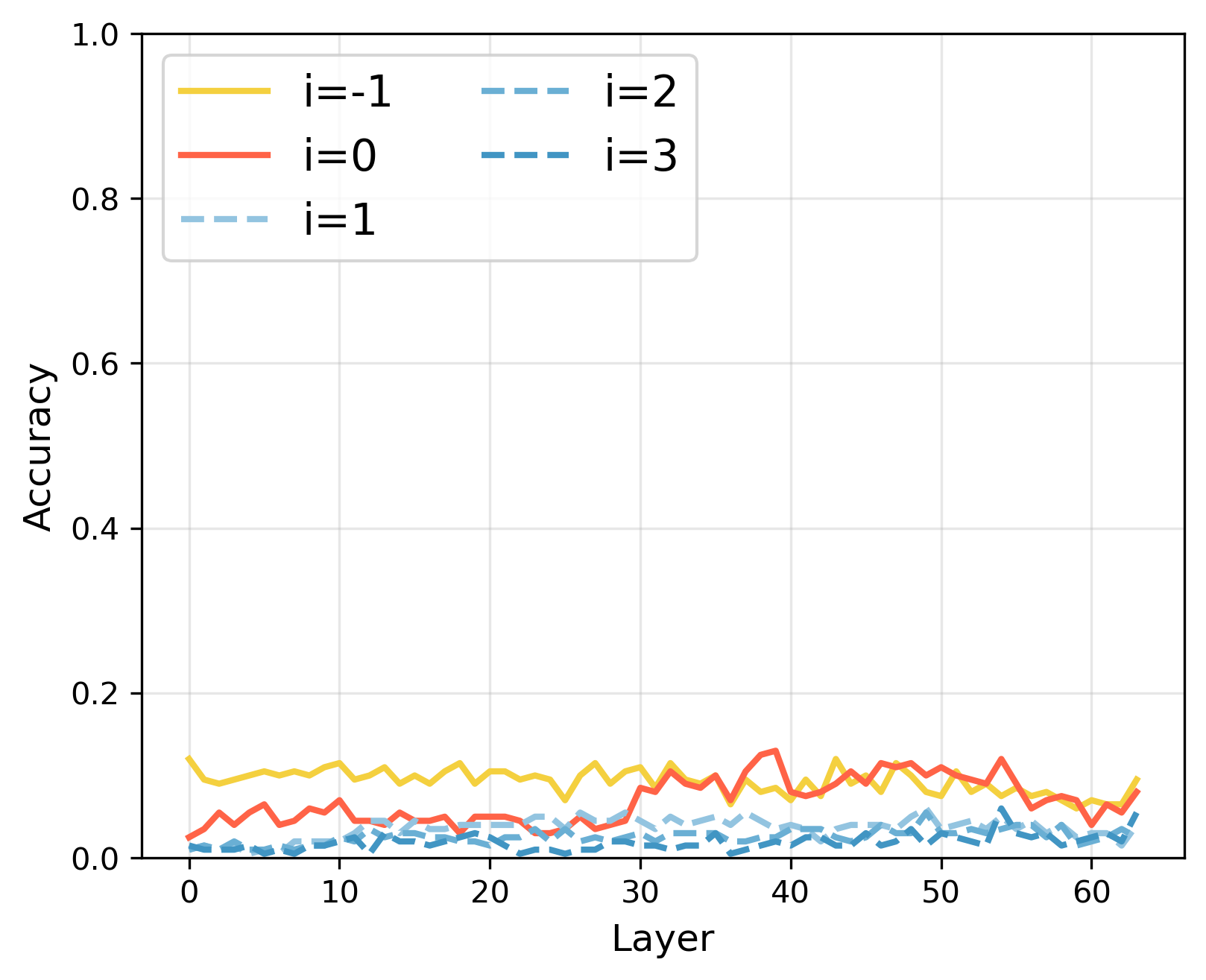}
    \end{subfigure}
    \begin{subfigure}{0.49\linewidth}
        \centering
        \subcaption[]{Gemma-3-27B}
        \includegraphics[width=\linewidth]{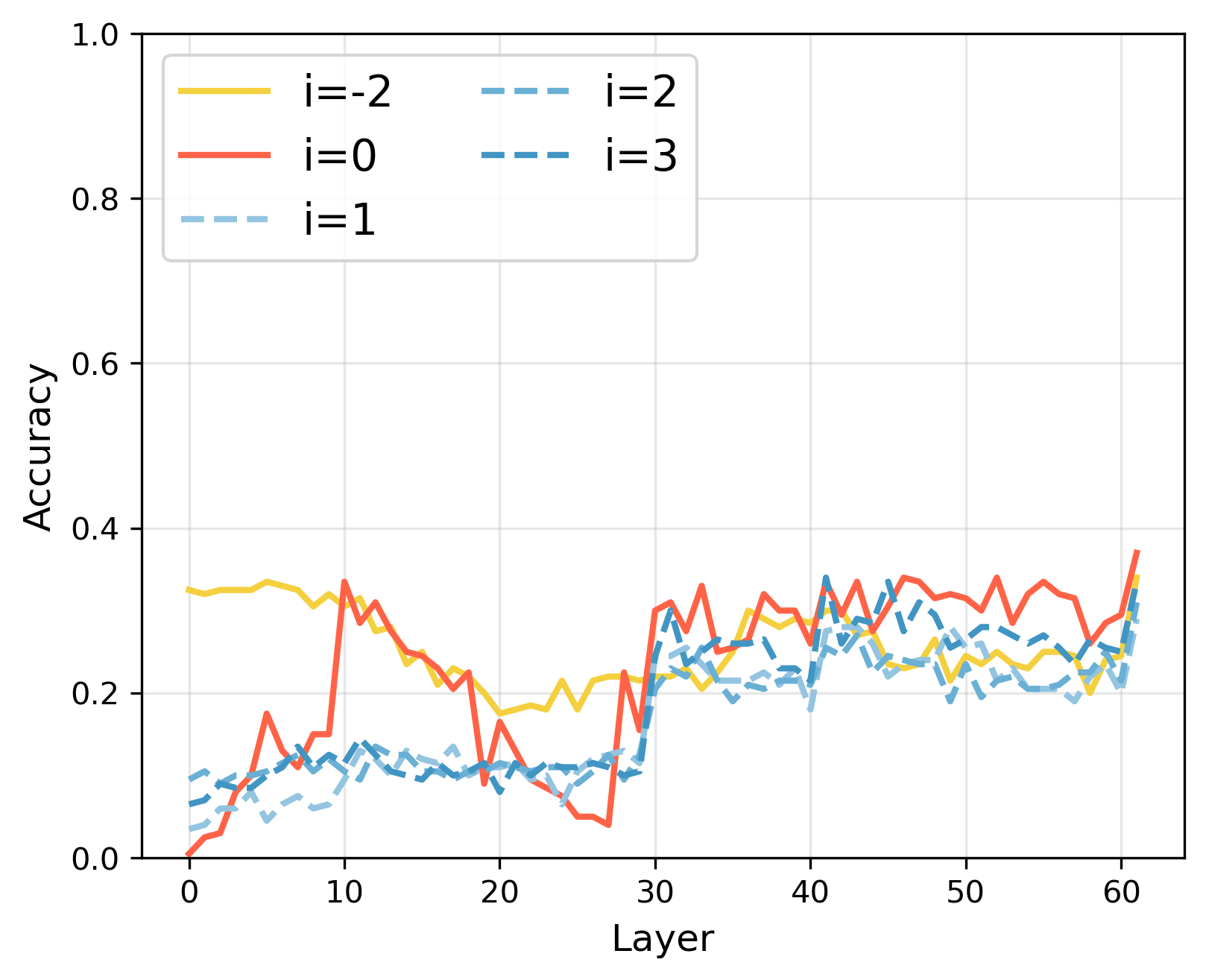}
    \end{subfigure}
    \begin{subfigure}{0.49\linewidth}
        \centering
        \subcaption[]{Llama-3.1-70B}
        \includegraphics[width=\linewidth]{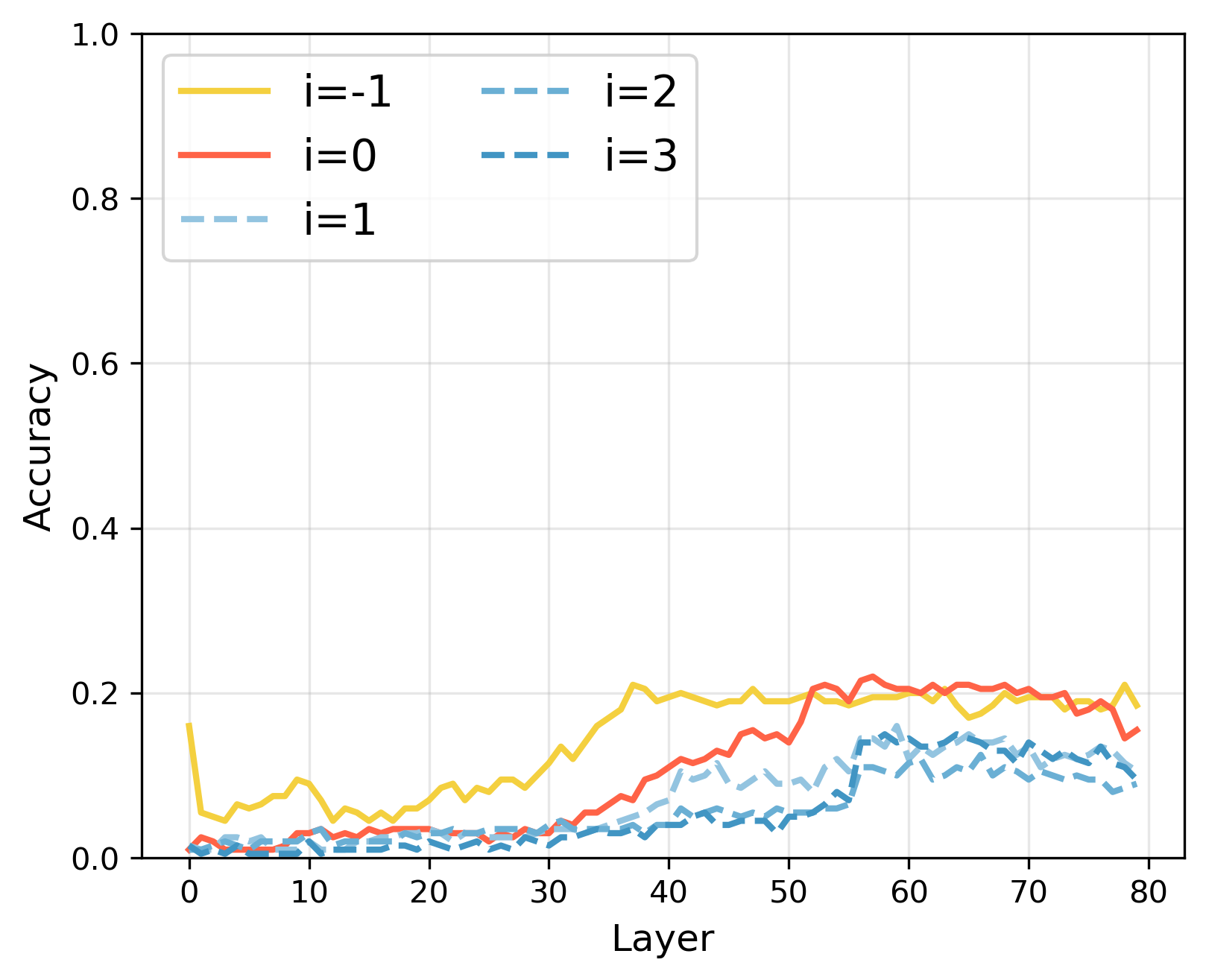}
    \end{subfigure}
    \caption{Top-1 probe accuracy predicting $r_2$ on rhyming couplets. Mirrors the top-5 and rhyme accuracy results: the $i\leq0$ probes show substantially higher accuracy in middle-to-late layers than probes at $i>0$.}
    \label{fig:appendix-poem}
\end{figure}

\section{Additional Activation Patching Details}
\label{sec:additional-patching-details}

\subsection*{Baselines}
\label{sec:patching-baselines}

To verify that the observed corrupt rhyme rates reflect the specific encoding of the corrupt rhyme word rather than a generic effect of perturbing the residual stream, we run two control conditions. The \textit{zero-vector} baseline replaces the patched hidden state with an all-zeros vector. The \textit{donor-prompt} baseline replaces it with a hidden state cached from the same token position in a semantically unrelated sentence (``The weather outside is warm and sunny today, and the birds are singing.'').

\begin{figure}[ht]
    \centering
    \begin{subfigure}{\linewidth}
        \centering
        \includegraphics[width=\linewidth]{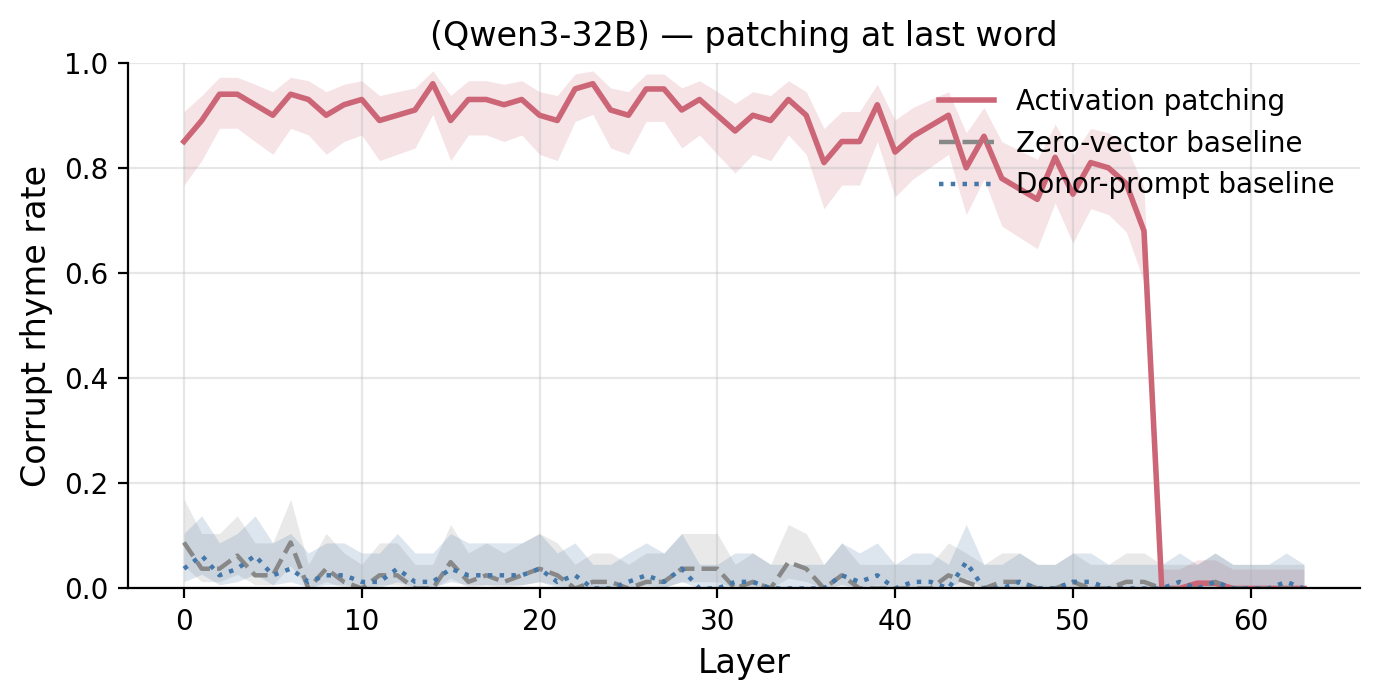}
    \end{subfigure}
    \begin{subfigure}{\linewidth}
        \centering
        \includegraphics[width=\linewidth]{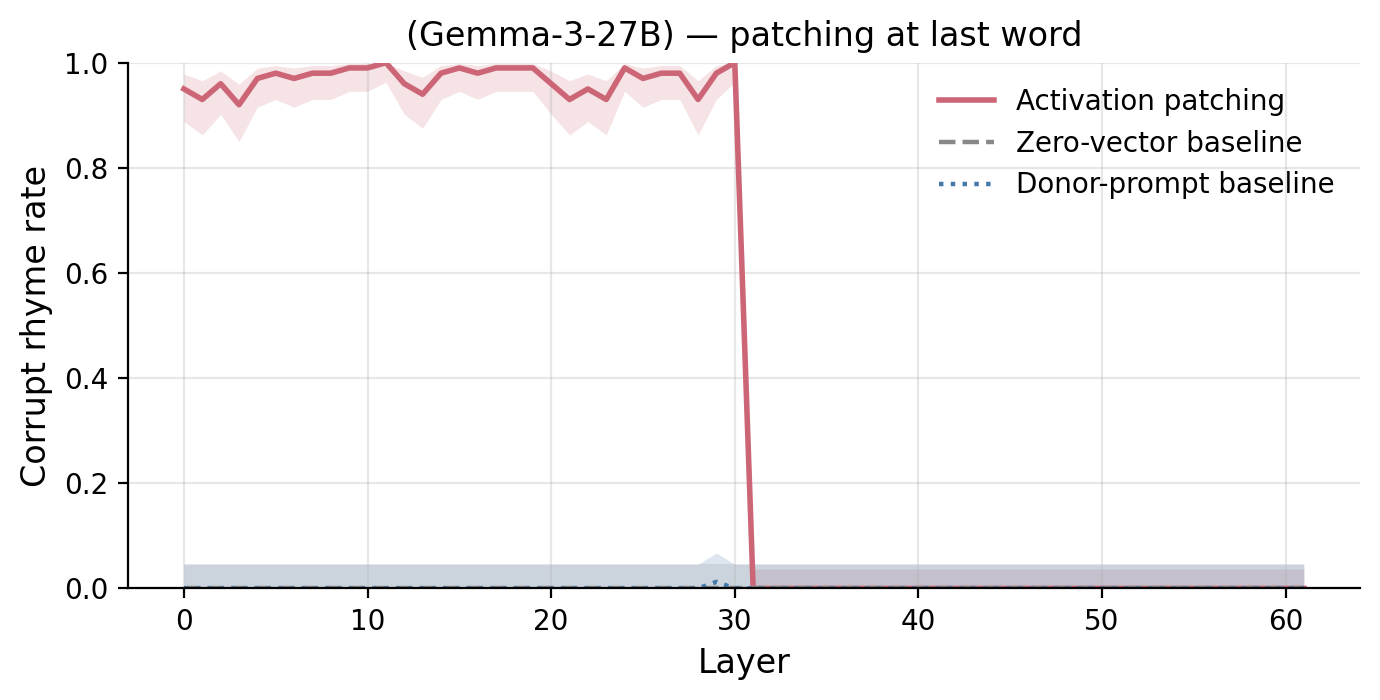}
    \end{subfigure}
    \caption{Corrupt rhyme rate under true activation patching versus zero-vector and donor-prompt baselines at the last word position. Shaded bands are Wilson 95\% CIs (true patching $N{=}100$ per layer; baselines pooled across pairs $N\approx 80$ per layer). Both baselines yield near-zero rates across all layers, confirming that the patching effect is specific to the corrupt rhyme word's identity rather than an artifact of residual stream disruption.}
    \label{fig:patching-baselines}
\end{figure}

Figure~\ref{fig:patching-baselines} shows results for Qwen3-32B (at $i=-1$) and Gemma-3-27B (at $i=-2$), averaged over all prompt pairs. Both control conditions produce corrupt rhyme rates near zero across all layers, while true activation patching reaches a peak of 0.96 for Gemma-3-27B and 0.90 for Qwen3-32B (single fright/fear prompt with $N{=}100$). Across all layers and both control conditions, the Wilson 95\% upper bound is at most 0.17, well below the true patching peak. This supports the claim that the patching effect is specific to the corrupt rhyme word's identity rather than a generic perturbation artifact.

\subsection*{All-Layers Patching}
\label{sec:patch-all-layers}

Table~\ref{tab:patch-all-layers} reports corrupt rhyme rates when all layers are patched simultaneously at a given token position. Positions $i \geq 1$ yield zero corrupt rhyme rate and are omitted.

\begin{table}[h]
\centering
\small
\begin{tabular}{llcc}
\toprule
Family & Model & Last Word [95\% CI] & $i=0$ [95\% CI] \\
\midrule
\multirow{6}{*}{Qwen3}
 & 0.6B  & 1 [0, 5]    & 7 [3, 14]  \\
 & 1.7B  & 15 [9, 23]  & 2 [1, 7]   \\
 & 4B    & 42 [33, 52] & 7 [3, 14]  \\
 & 8B    & 53 [43, 62] & 2 [1, 7]   \\
 & 14B   & 63 [53, 72] & 7 [3, 14]  \\
 & 32B   & 76 [67, 83] & 1 [0, 5]   \\
\midrule
\multirow{4}{*}{Gemma-3}
 & 1B    & 37 [28, 47] & 1 [0, 5]   \\
 & 4B    & 78 [69, 85] & 0 [0, 4]   \\
 & 12B   & 90 [83, 94] & 0 [0, 4]   \\
 & 27B   & 85 [77, 91] & 67 [57, 75] \\
\midrule
\multirow{4}{*}{Llama-3}
 & 1B    & 59 [49, 68] & 2 [1, 7]   \\
 & 3B    & 87 [79, 92] & 0 [0, 4]   \\
 & 8B    & 79 [70, 86] & 0 [0, 4]   \\
 & 70B   & 75 [66, 82] & 2 [1, 7]   \\
\bottomrule
\end{tabular}
\caption{Corrupt rhyme rate (\%) when all layers are patched simultaneously, with Wilson 95\% CIs. All cells use $N{=}100$ (5 prompt pairs $\times\ N{=}20$). Gemma-3-27B is the only model whose newline-patching CI does not overlap zero, separated by a wide margin from every other model.}
\label{tab:patch-all-layers}
\end{table}

\subsection*{Per-Layer Results: All Models}

\begin{figure}[ht]
    \centering
    \includegraphics[width=\linewidth]{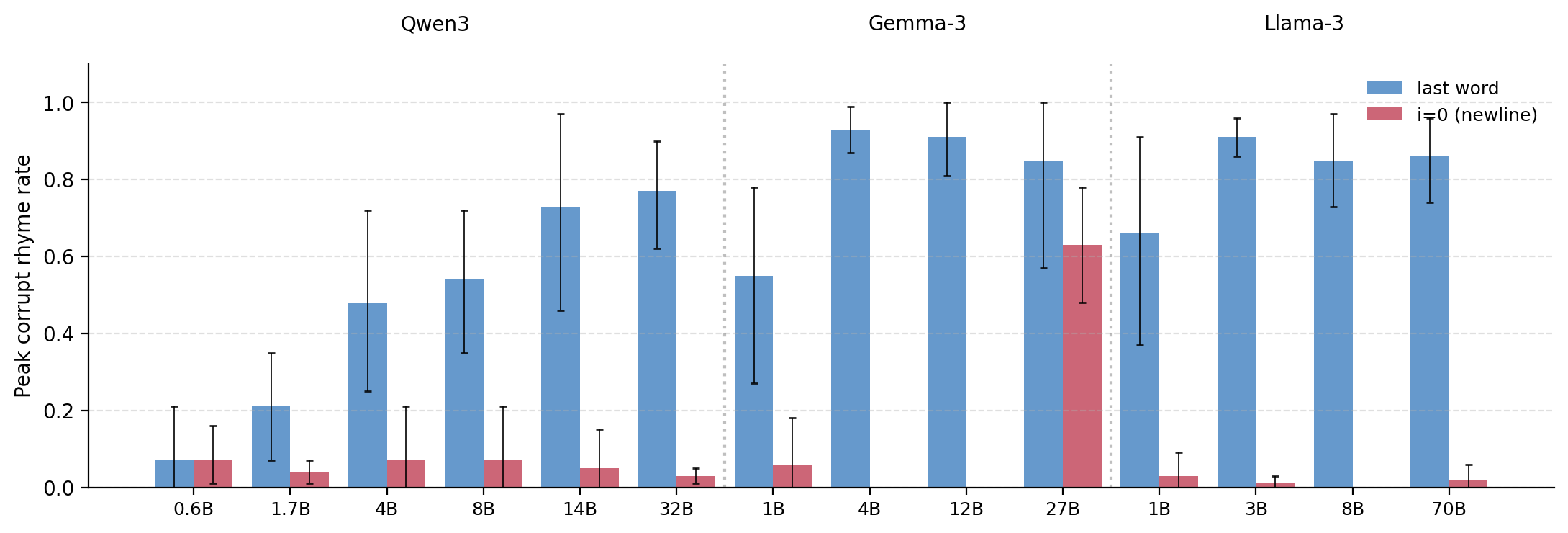}
    \caption{Peak corrupt rhyme rate (maximum across all layers) for the last word token and newline ($i=0$) at each model size. Black error bars are 95\% cluster bootstrap CIs over prompt pairs (5 pairs per model). Gemma-3-27B is the only model whose newline CI is clearly separated from zero (0.63 [0.48, 0.78]); every other model has a newline CI upper bound $\leq 0.21$ regardless of scale, while last-word patching is broadly effective from $\sim$3B parameters onward in every family.}
    \label{fig:appendix-patch-summary}
\end{figure}

\begin{figure}[ht]
    \centering
    \includegraphics[width=\linewidth]{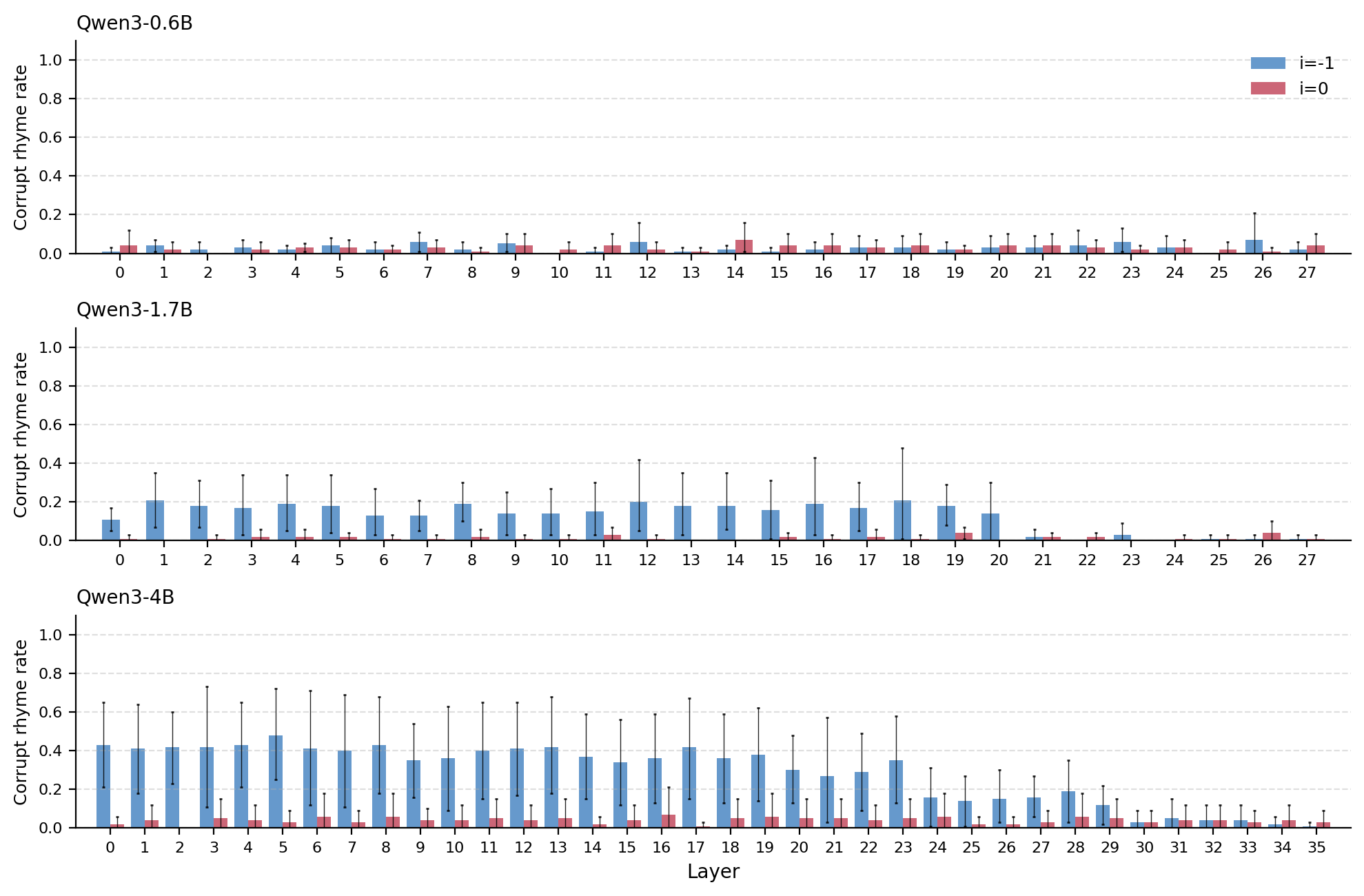}
    \caption{Per-layer activation patching, Qwen3 0.6B--4B. Black bars are 95\% cluster bootstrap CIs (5 pairs $\times\ N{=}20$). Last-word patching becomes effective only from 4B (peak 0.48 [0.25, 0.72]); 0.6B and 1.7B have peak CIs that nearly span zero. Newline ($i=0$) patching is at noise across all sizes (peak CI upper bounds $\leq 0.21$).}
    \label{fig:appendix-patch-qwen3-small}
\end{figure}

\begin{figure}[ht]
    \centering
    \includegraphics[width=\linewidth]{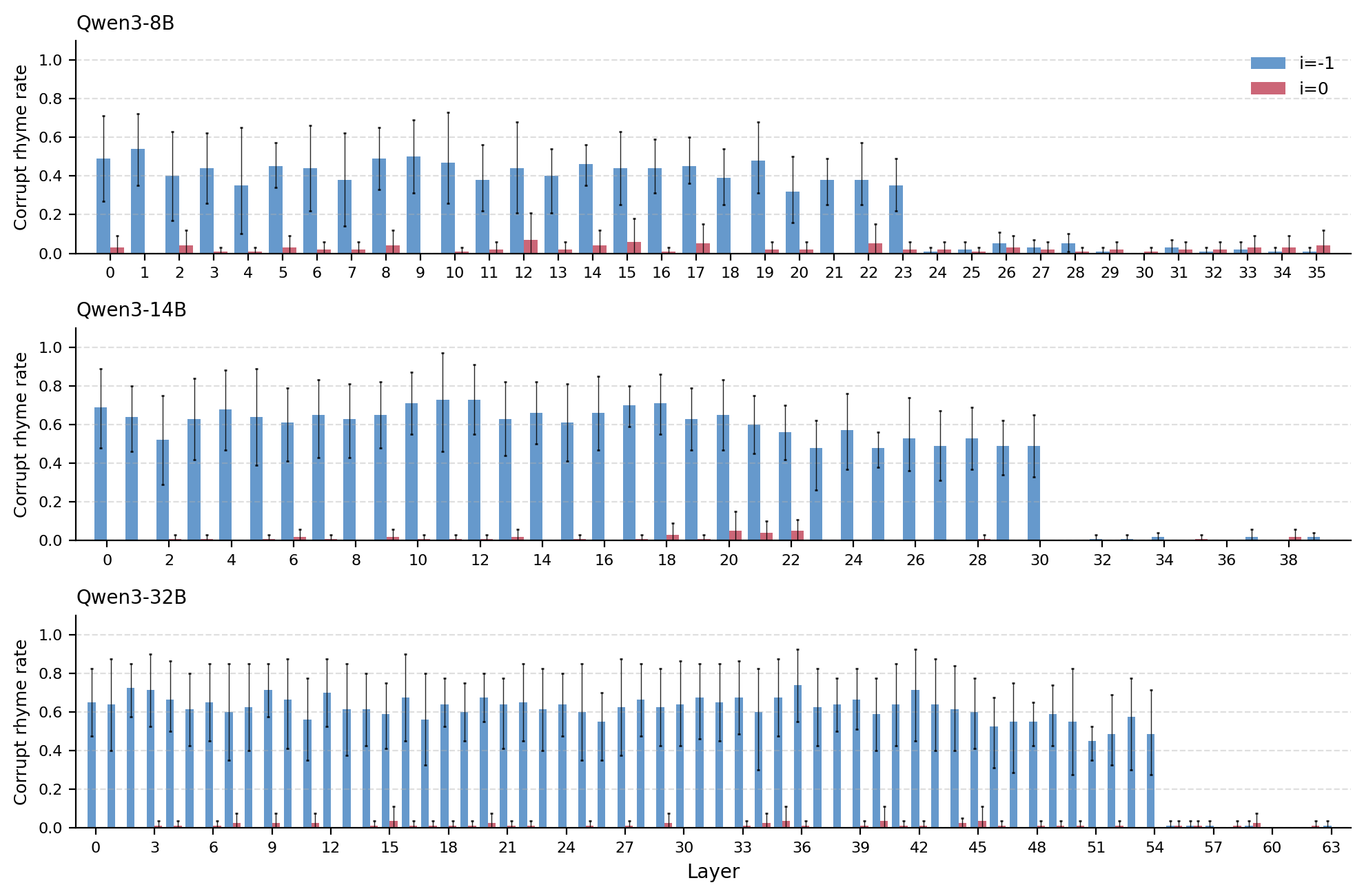}
    \caption{Per-layer activation patching, Qwen3 8B--32B. Black bars are 95\% cluster bootstrap CIs. Last-word peaks rise smoothly with scale (8B 0.54 [0.35, 0.72]; 14B 0.73 [0.46, 0.97]; 32B 0.74 [0.55, 0.93]). Newline patching remains at noise across all three sizes (CI upper bounds $\leq 0.21$).}
    \label{fig:appendix-patch-qwen3-large}
\end{figure}

\begin{figure}[ht]
    \centering
    \includegraphics[width=\linewidth]{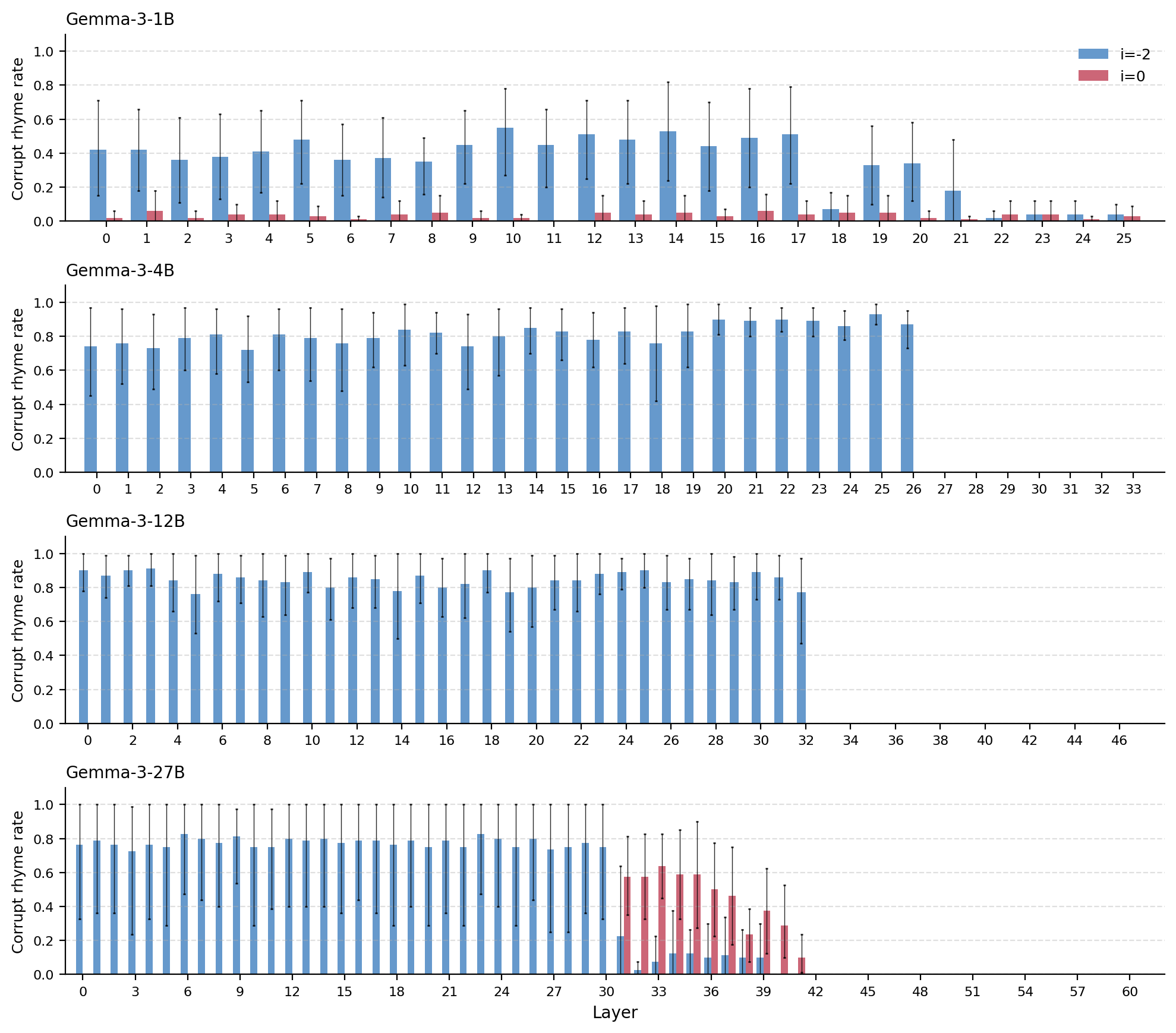}
    \caption{Per-layer activation patching, Gemma-3 1B--27B. Black bars are 95\% cluster bootstrap CIs. The newline ($i=0$) channel is silent at every size below 27B (peak CI [0.00, 0.00] for 4B and 12B), and only emerges at 27B with peak 0.63 [0.48, 0.78] at L33. Last-word ($i=-2$) patching is effective from 1B onward.}
    \label{fig:appendix-patch-gemma3}
\end{figure}

\begin{figure}[ht]
    \centering
    \includegraphics[width=\linewidth]{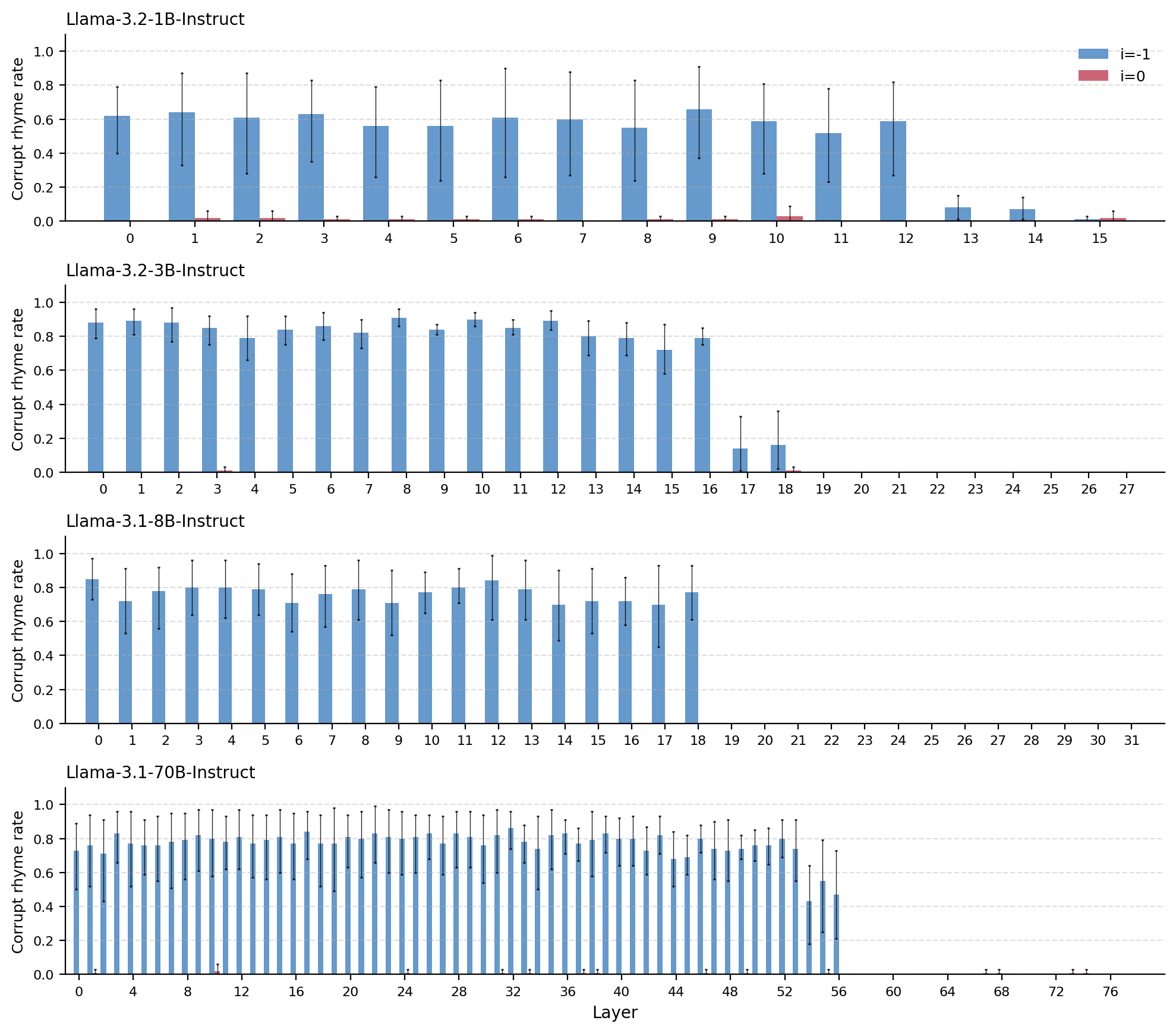}
    \caption{Per-layer activation patching, Llama-3 1B--70B. Black bars are 95\% cluster bootstrap CIs. Last-word peaks are high from 3B onward (3B 0.91 [0.86, 0.96]; 8B 0.85 [0.73, 0.97]; 70B 0.86 [0.74, 0.96]); 1B is lower (peak 0.66 [0.56, 0.75]). Newline patching is at noise across all four sizes (CI upper bounds $\leq 0.09$).}
    \label{fig:appendix-patch-llama3}
\end{figure}

\subsection*{Full Position Sweep}
\label{sec:additional-patching-sweep}

\begin{figure}[ht]
    \centering
    \begin{subfigure}{0.49\linewidth}
        \centering
        \subcaption[]{Qwen3-32B}
        \includegraphics[width=\linewidth]{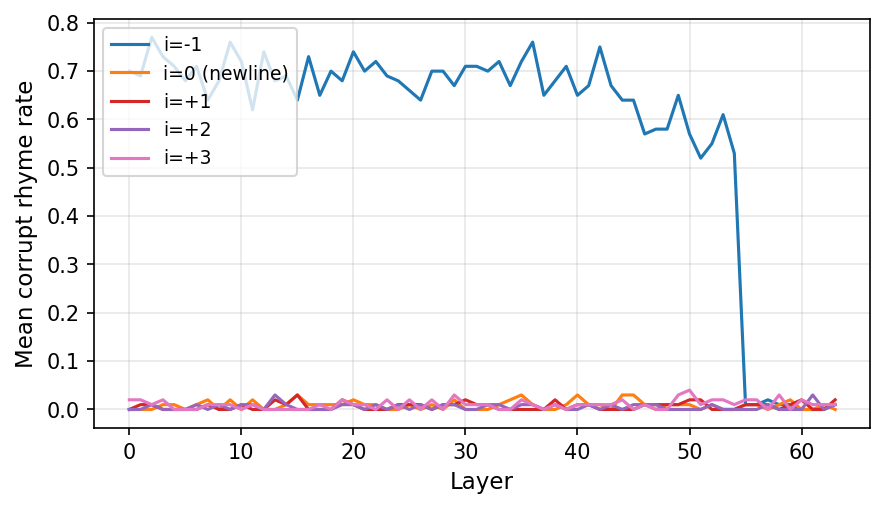}
    \end{subfigure}
    \begin{subfigure}{0.49\linewidth}
        \centering
        \subcaption[]{Gemma-3-27B}
        \includegraphics[width=\linewidth]{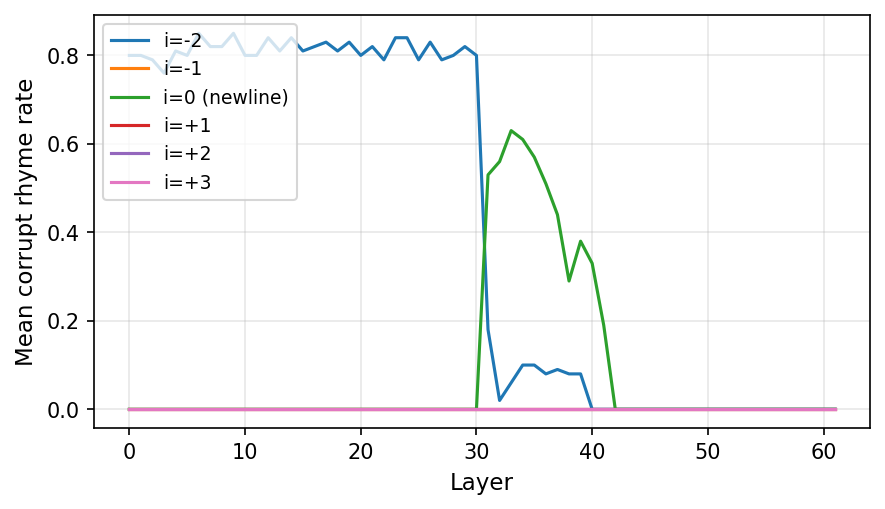}
    \end{subfigure}
    \caption{Corrupt rhyme rate across all six swept token positions, averaged over 5 prompt pairs. For Gemma-3-27B, the comma token at $i=-1$ is near zero throughout, confirming the handoff is specific to the newline token.}
    \label{fig:patching-all-positions}
\end{figure}

\begin{figure*}[ht]
    \centering
    \begin{subfigure}{0.49\linewidth}
        \centering
        \includegraphics[width=\linewidth]{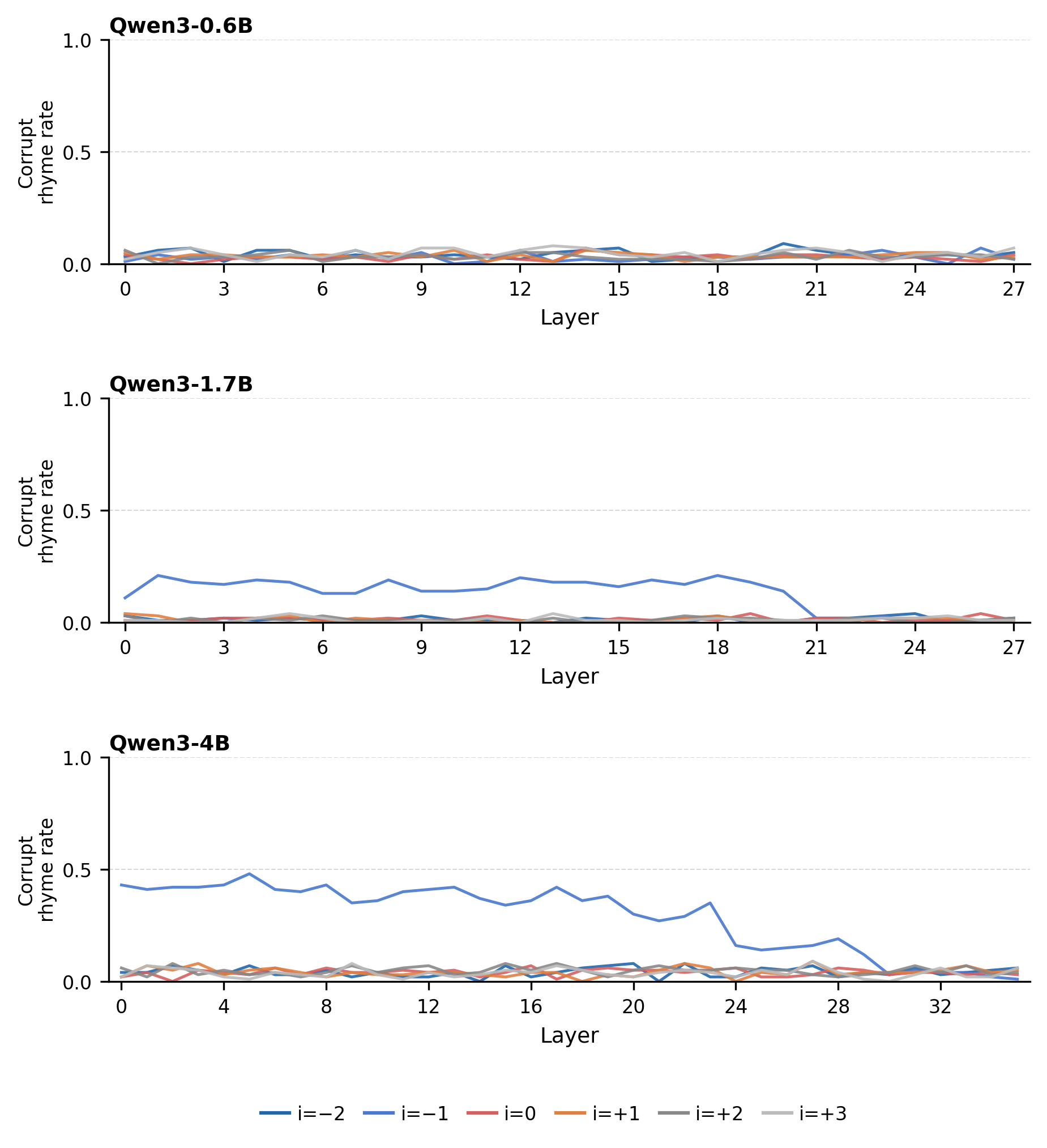}
        \subcaption{Qwen3 0.6B--4B.}
        \label{fig:appendix-sweep-qwen3-small}
    \end{subfigure}
    \hfill
    \begin{subfigure}{0.49\linewidth}
        \centering
        \includegraphics[width=\linewidth]{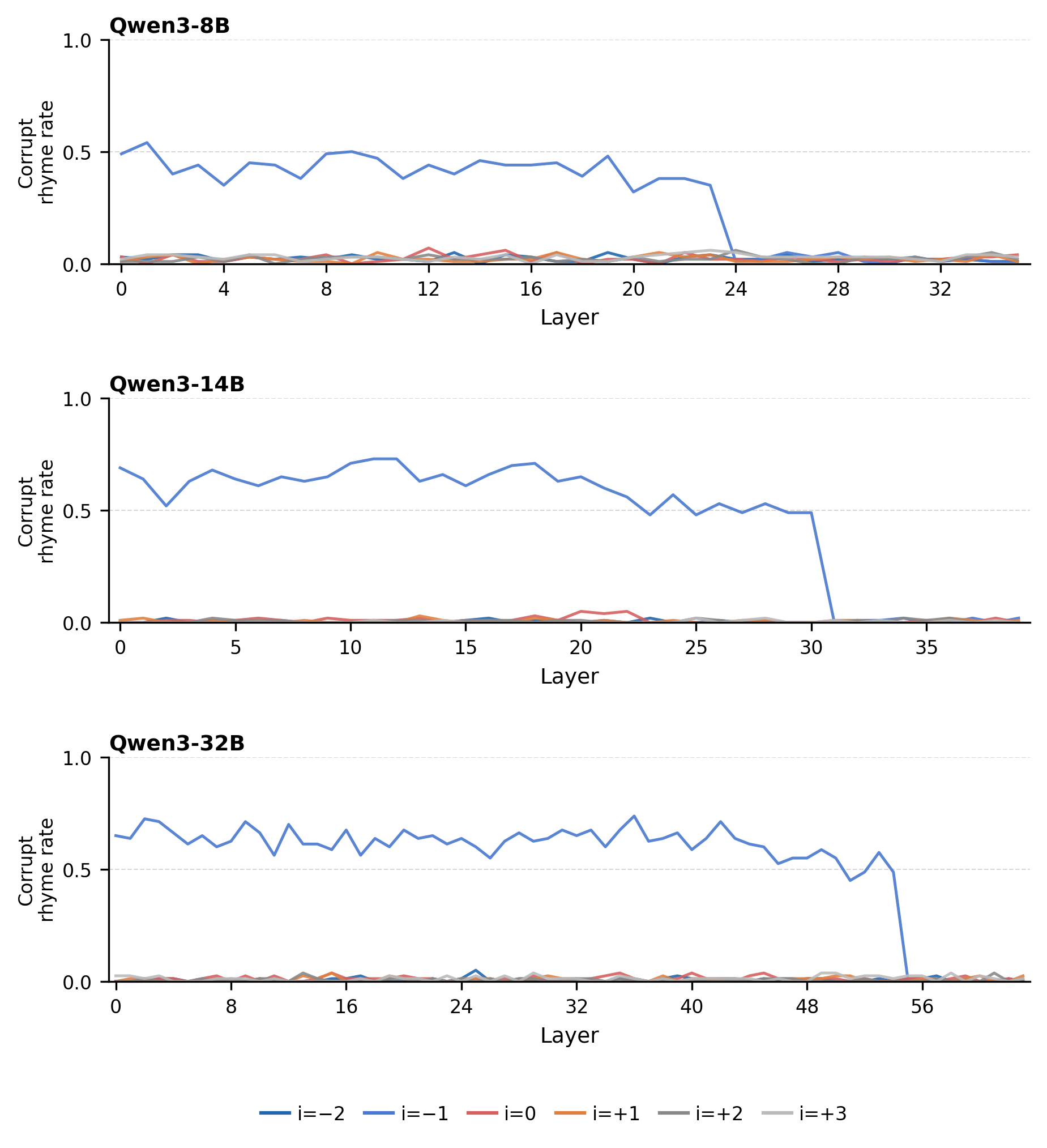}
        \subcaption{Qwen3 8B--32B.}
        \label{fig:appendix-sweep-qwen3-large}
    \end{subfigure}

    \medskip

    \begin{subfigure}{0.49\linewidth}
        \centering
        \includegraphics[width=\linewidth]{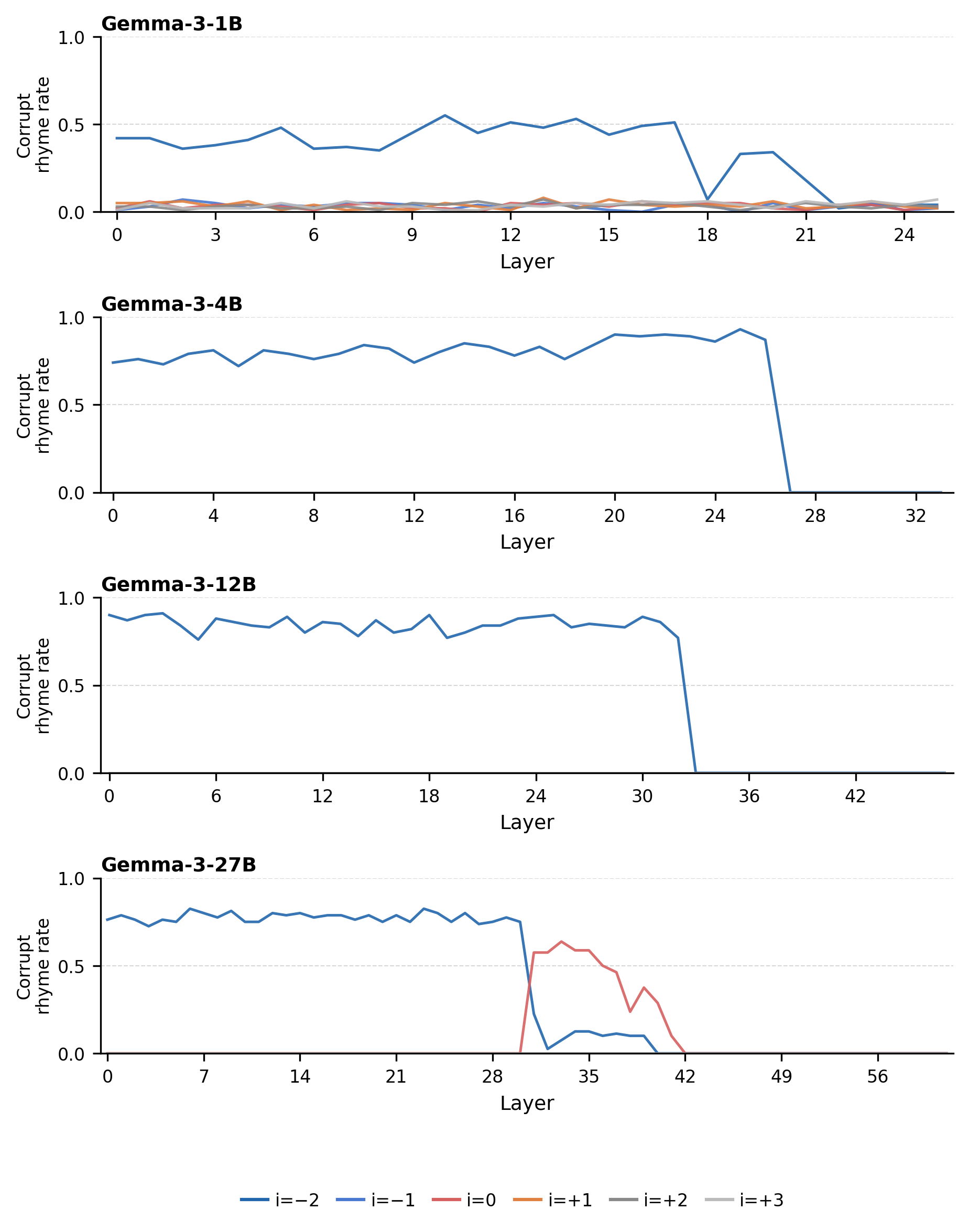}
        \subcaption{Gemma-3 1B--27B.}
        \label{fig:appendix-sweep-gemma3}
    \end{subfigure}
    \hfill
    \begin{subfigure}{0.49\linewidth}
        \centering
        \includegraphics[width=\linewidth]{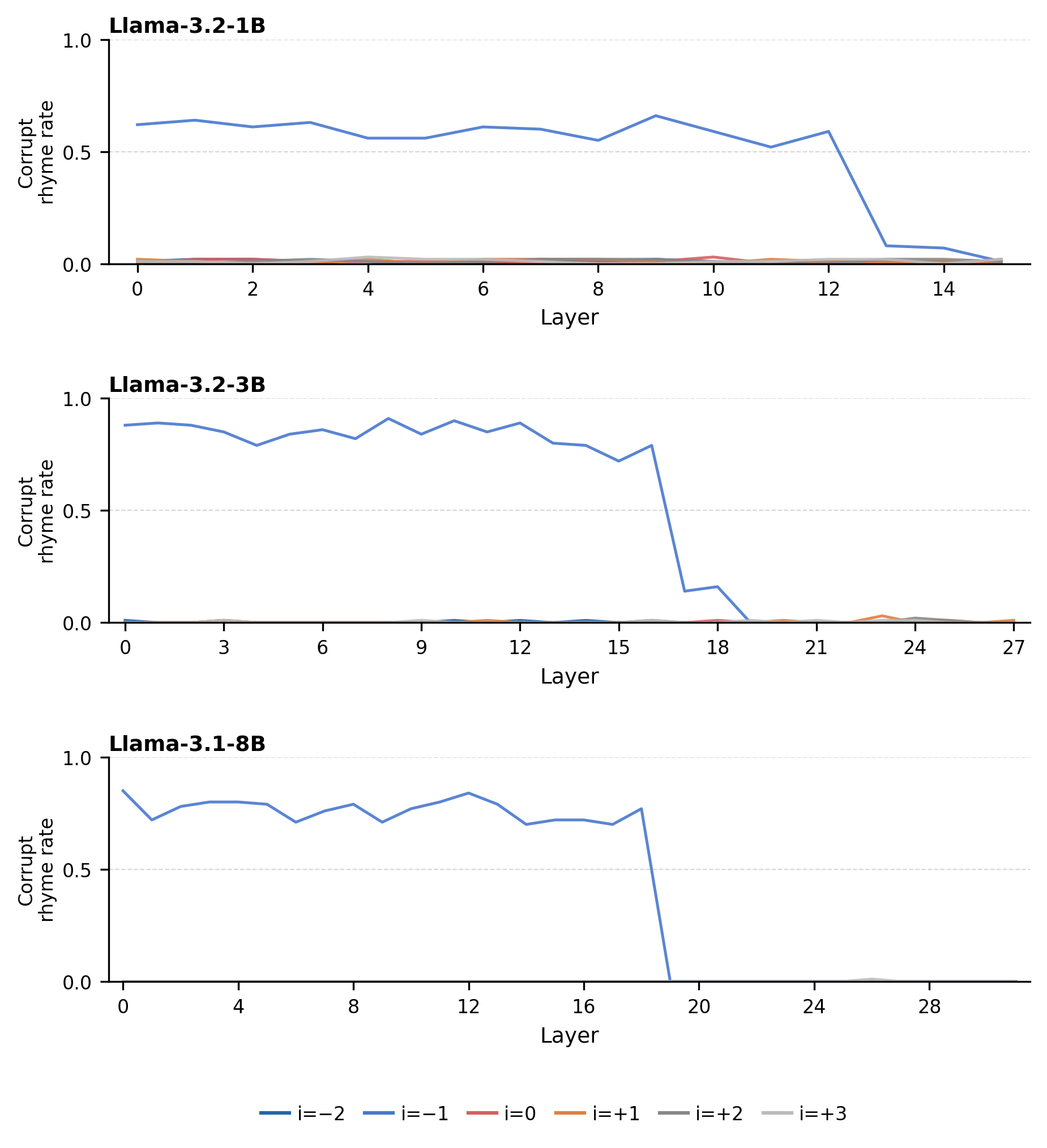}
        \subcaption{Llama-3 1B--70B.}
        \label{fig:appendix-sweep-llama3}
    \end{subfigure}
    \caption{Full position sweep across all four model groups in the Qwen3, Gemma-3, and Llama-3 families.}
    \label{fig:appendix-sweep}
\end{figure*}

\subsection*{Prompt Pairs}

For each pair, we show the clean prompt, the corrupt prompt, and an example patched completion
where activation patching was successful in steering the rhyme scheme.

{\small\raggedright
\medskip\noindent\textbf{Pair 1: doom/dread, $\ell=10$, $i=-1$}\\
\textit{Clean:} \texttt{...filled with silent doom,\textbackslash n when suddenly they}\\
\textit{Corrupt:} \texttt{...filled with silent dread,\textbackslash n when suddenly they}\\
\textit{Patched:} \texttt{when suddenly they heard a creaking bed.}

\medskip\noindent\textbf{Pair 2: bliss/joy, $\ell=1$, $i=-2$}\\
\textit{Clean:} \texttt{...The children laughed in bliss,\textbackslash n until they all}\\
\textit{Corrupt:} \texttt{...The children laughed in joy,\textbackslash n until they all}\\
\textit{Patched:} \texttt{until they all became a toy.}

\medskip\noindent\textbf{Pair 3: dark/night, $\ell=0$, $i=-2$}\\
\textit{Clean:} \texttt{...She wandered home alone into the dark,\textbackslash n and then she}\\
\textit{Corrupt:} \texttt{...She wandered home alone into the night,\textbackslash n and then she}\\
\textit{Patched:} \texttt{and then she saw a strange and eerie light.}

\medskip\noindent\textbf{Pair 4: grief/pain, $\ell=2$, $i=-1$}\\
\textit{Clean:} \texttt{...I never knew the depth of such grief,\textbackslash n as though the}\\
\textit{Corrupt:} \texttt{...I never knew the depth of such pain,\textbackslash n as though the}\\
\textit{Patched:} \texttt{as though the sky had lost its rain.}

\medskip\noindent\textbf{Pair 5: fright/fear, $\ell=1$, $i=-1$}\\
\textit{Clean:} \texttt{...She felt a sudden sense of fright,\textbackslash n and hoped that}\\
\textit{Corrupt:} \texttt{...She felt a sudden sense of fear,\textbackslash n and hoped that}\\
\textit{Patched:} \texttt{and hoped that someone would appear.}
}

\end{document}